\documentclass{article}
\usepackage{iclr2027_conference,times}

\usepackage{amsmath,amsfonts,bm}

\def\eqref#1{equation~\ref{#1}}

\def\1{\bm{1}}

\DeclareMathAlphabet{\mathsfit}{\encodingdefault}{\sfdefault}{m}{sl}
\SetMathAlphabet{\mathsfit}{bold}{\encodingdefault}{\sfdefault}{bx}{n}

\usepackage{hyperref}
\usepackage{url}
\usepackage{xcolor}
\usepackage{booktabs}
\usepackage{array}
\usepackage{colortbl}
\definecolor{oursrow}{HTML}{E4EEFB}
\usepackage{graphicx}
\usepackage{wrapfig}
\usepackage{xspace}
\usepackage{fontawesome5}
\definecolor{linkblue}{HTML}{2F5DA8}
\newcommand{\plink}[2]{{\hypersetup{pdfborder={0 0 0}}\href{#1}{\textcolor{linkblue}{#2}}}}

\newcommand{\method}{{LeapQuant}}
\newcommand{\sys}{\method\xspace}

\title{\method{}: Efficient Linear Attention with Accurate Recurrent State Quantization
}

\author{
\begin{minipage}[t]{0.96\textwidth}
Yi Pan$^1$\thanks{Equal contribution}\quad Haocheng Xi$^1$\footnotemark[1]\quad Kan Zhu$^2$\quad Xingyang Li$^3$\quad Yibo Wu$^4$\\[0.2em]
Mayank Mishra$^1$\quad Hongtao Zhang$^2$\quad William X. Zheng$^1$\quad Baris Kasikci$^2$\\[0.2em]
Song Han$^{3,5}$\quad Kurt Keutzer$^1$\quad Rishabh Iyer$^1$\quad Ion Stoica$^1$ \\
\end{minipage}\\
$^1$UC Berkeley, $^2$University of Washington,  $^3$MIT, $^4$Perplexity AI, $^5$NVIDIA\\[0.5em]
\plink{https://leapquant.github.io}{\faGlobe~\textbf{Project Page}}\qquad \plink{https://github.com/leapquant/LeapQuant}{\faGithub~\textbf{Code}}
}

\newcommand{\hlinline}[2]{$#2$}
\newcommand{\hfp}[1]{#1}
\newcommand{\hq}[1]{#1}
\newcommand{\hrec}[1]{#1}

\DeclareMathOperator{\Diag}{Diag}
\newcommand{\quant}{\operatorname{\mathtt{quant}}}
\newcommand{\dequant}{\operatorname{\mathtt{dequant}}}
\newcommand{\DQ}{\dequant}

\newcommand{\Sh}{\hat{S}}

\AddToHook{env/table/begin}{\setlength{\belowcaptionskip}{4pt}}
\AddToHook{env/wraptable/begin}{\setlength{\belowcaptionskip}{4pt}}
\makeatletter
\def\paragraph{\@startsection{paragraph}{4}{\z@}
  {0.75ex plus 0.25ex minus .1ex}
  {-1em}
  {\normalsize\bf}}
\makeatother

\iclrfinalcopy
\arxivtrue

\begin{document}

\maketitle

\begin{abstract}

Recent LLMs increasingly adopt hybrid designs that replace standard attention with linear attention, such as Gated DeltaNet (GDN) and Kimi Delta Attention (KDA).
Although they compress the context into a fixed-size recurrent state and substantially reduce the cost of long-context processing, repeatedly reading and updating that state remains a major inference bottleneck.
Quantization offers a natural way to reduce this cost, but can significantly degrade model quality, due to the accumulation of rounding errors and the presence of outlier rows and columns in the state.
To address these challenges, we propose \emph{\method{}}, a training-free method that achieves near-lossless performance under 8-bit recurrent-state quantization.
First, to mitigate error accumulation, we propose per-window quantization, which \emph{leaps} over a window of tokens and quantizes the state only once at its end.
Within a window, outputs are computed from the fixed low-bit state together with high-precision buffered updates.
Second, to reduce the error introduced by each quantization, \method{} retains the state's largest outliers as a few high-precision \emph{Compensator Tokens}, which share the update path of real tokens.
We then smooth the remaining residual before quantization to further reduce the error.
Comprehensive experiments across the Qwen, Kimi, and GLM model families show that \emph{\method{}} substantially reduces memory and compute costs during inference.
With accuracy comparable to the FP32 baseline, it achieves average speedups of 2.05--3.70$\times$ at the kernel level and 1.51$\times$ for end-to-end inference on NVIDIA B200, RTX PRO 6000, and RTX 5090 GPUs.

\end{abstract}

\section{Introduction}\label{sec:introduction}

Linear attention, which compresses the token history into a fixed-size recurrent state, is now widely used in long-context large language models.
For instance, recent hybrid architectures in the Qwen, Kimi, and GLM model families replace many standard attention layers with recurrent linear attention layers such as Gated DeltaNet (GDN) and Kimi Delta Attention (KDA)~\citep{qwen2025qwen3next,yang2025gated,team2025kimi}.
By compressing the token history, these layers reduce the computation and memory growth associated with long context, making inference more efficient.

\begin{figure}[t]
    \centering
    \includegraphics[width=\linewidth]{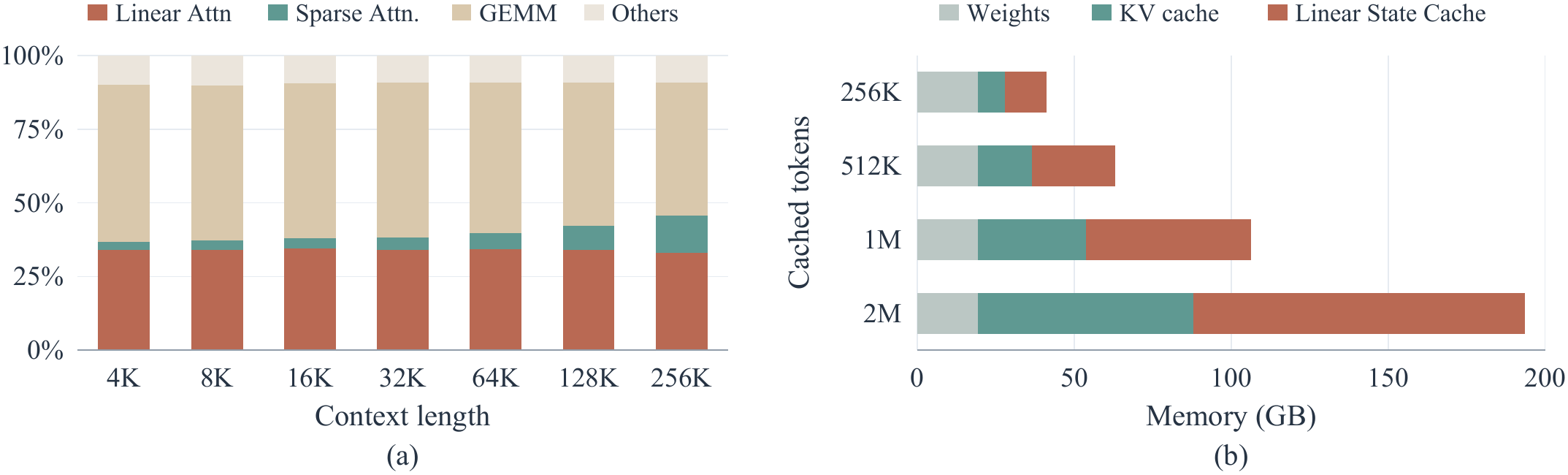}
    \caption{Recurrent-state access is a major cost of LLM serving.
    (a) Decode-time breakdown for GLM-5.3-Flash-NVFP4 on the B200 GPU at batch size 256. The share spent on linear attention stays roughly constant as context length grows.
    (b) GPU memory footprint for Qwen3.5-9B with prefix caching, assuming one linear attention state per 1{,}024 cached tokens.}
    \vspace{-1em}
    \label{fig:intro}
\end{figure}

However, the recurrent state in linear attention still imposes substantial memory bandwidth and capacity costs during inference.
For each generated token, a linear-attention layer loads its state matrix from HBM, applies a lightweight update, computes the token output, and writes the updated matrix back to HBM.
This full-state transfer repeats at every decoding step.
Since the update and readout perform little arithmetic relative to the bytes transferred, inference throughput is bottlenecked by HBM bandwidth~\citep{williams2009roofline}.
Consequently, reading and writing the recurrent states across many linear attention layers and concurrent requests accounts for a significant fraction of decoding time, as shown in Figure~\ref{fig:intro}(a).
Additionally, the recurrent state consumes substantial GPU memory when prefix caching is enabled, as serving systems retain separate states for each cached prefix (Figure~\ref{fig:intro}(b))~\citep{pan2025marconi}.

Quantizing the recurrent state is a natural way to reduce both memory traffic and footprint, but doing so in a way that preserves model accuracy is challenging.
In particular, a naive approach that stores the state in a standard low-bit format and re-quantizes it after every token can substantially degrade accuracy, especially over long thinking traces.
We find that accuracy degradation arises from two sources of error.
First, quantization error accumulates recurrently: each update starts from an already quantized state, and quantizing the result introduces another rounding error, causing the state to deviate progressively from the FP32 trajectory over a long generation (Figure~\ref{fig:state-error}).
Second, large outliers are often concentrated in a few rows and columns of the state.
These outliers widen the quantization range and force smaller values onto coarse quantization levels, amplifying the error introduced each time the state is quantized.
Accurate low-bit quantization therefore requires reducing both how often error is introduced and how much error each quantization introduces.

\begin{figure}[t]
    \centering
    \includegraphics[width=0.9\linewidth]{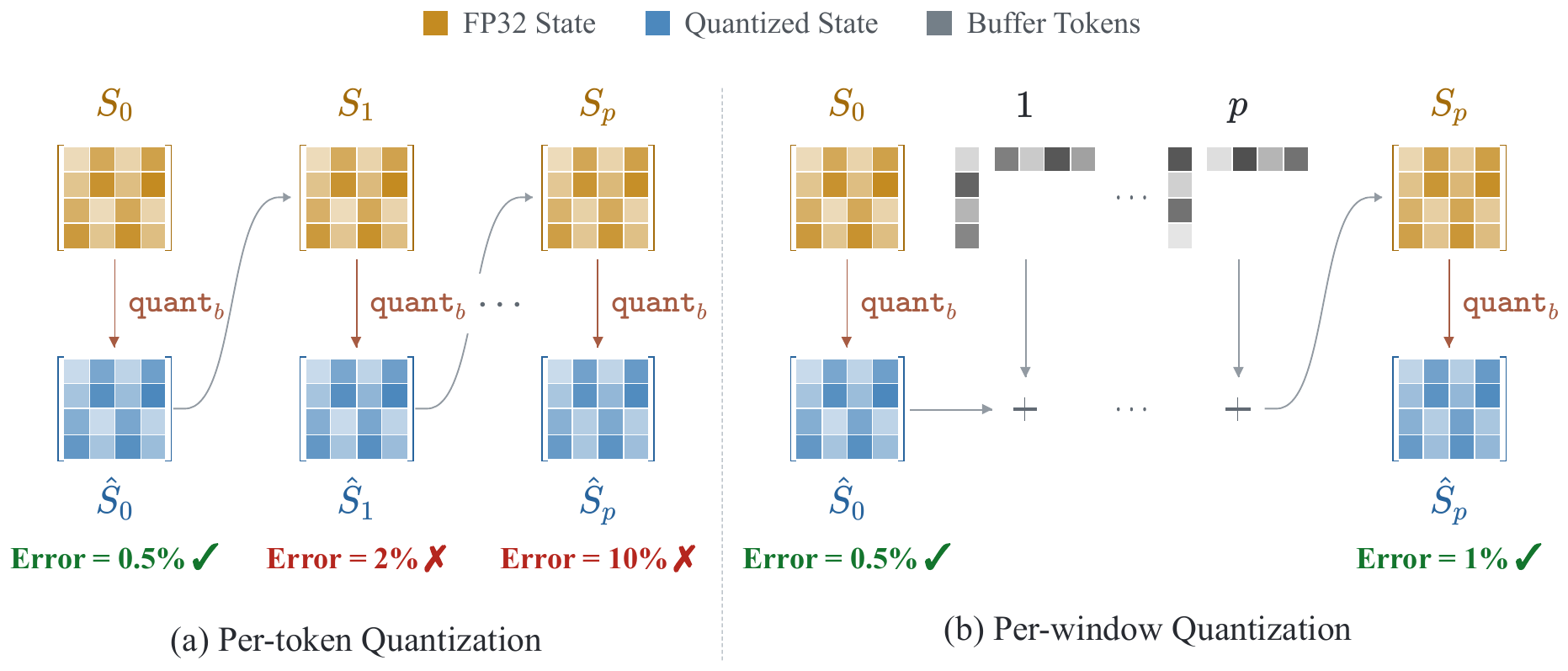}
    \caption{Per-token versus per-window quantization. Per-token quantization re-quantizes the full state after every update. Per-window quantization holds the quantized boundary state $\hat S_0$ fixed, buffers the $p$ updates of the window in higher precision, and reconstructs the state from them; the full state is quantized only once per window, into $\hat S_p$.}
    \label{fig:per-window}
    \vspace{-1.5em}
\end{figure}

We propose \emph{\sys{}}, a training-free method for recurrent-state quantization that achieves near-lossless model quality.
\sys{} introduces two key ideas that directly address the above sources of error.
First, \emph{per-window quantization} limits error accumulation by allowing \sys{} to \emph{leap} over a window of tokens before quantizing the state again (Figure~\ref{fig:per-window}).
At the beginning of each window, \sys{} stores the recurrent state in low precision.
Within the window, it holds this state fixed, buffers higher-precision token updates, and computes each output from the fixed state and buffered updates.
Only at the end of the window does it reconstruct and quantize the full updated state for the next window.
As a result, \sys{} introduces quantization error less frequently, slowing its accumulation over long contexts.

To address quantization error caused by outliers, \sys{} introduces \emph{Compensator Tokens}, which capture the state's largest outliers in high precision and thereby reduce the error of each quantization (Figure~\ref{fig:ghost-tokens}).
Each Compensator Token represents a high-precision outer product of two vectors, capturing large-magnitude patterns across both rows and columns and leaving a residual that is easier to quantize.
Since these rank-one terms have the same form as real-token updates, \sys{} incorporates them directly into the window's update path using only a few additional vectors and no separate recurrent state update.
\sys{} further smooths the residual across channels before quantization to reduce error.
Both key techniques in \sys{} are training-free and require no calibration data.

Our evaluation spanning the Qwen, Kimi, and GLM families demonstrates that \emph{\sys{}} substantially reduces memory costs and improves inference throughput.
Across 12 model--task pairs, \sys{} achieves accuracy comparable to the FP32 baseline while reducing state memory traffic by 3.4$\times$ and end-to-end memory footprint by up to 56\%.
Across NVIDIA B200, RTX PRO 6000, and RTX 5090 GPUs, \sys{} achieves average speedups of 2.05--3.70$\times$ at the kernel level and 1.51$\times$ for end-to-end inference.

\section{Related Work}

\paragraph{Linear attention and hybrid models.}
\cite{katharopoulos2020transformers} propose Linear attention to replace the softmax attention~\citep{vaswani2017attention} with a recurrence over a fixed-size state, and a series of works improve its expressiveness with data-dependent gating and delta-rule updates, including GLA~\citep{yang2023gated}, Mamba2~\citep{dao2024transformers}, DeltaNet~\citep{schlag2021linear,yang2024parallelizing}, Gated DeltaNet~\citep{yang2025gated}, and KDA~\citep{team2025kimi}.
Their chunkwise formulation~\citep{yang2023gated} materializes the state in HBM only at chunk boundaries during training and prefill, and is implemented using efficient Triton kernels~\citep{triton} in libraries such as FLA~\citep{yang2024fla}.
Recent LLMs combine a few full or sparse attention layers with a majority of linear attention layers~\citep{qwen2025qwen3next,team2026kimi,zai2026glm53flash,blakeman2025nemotron}, and these hybrid models have become a common design for long-context and long-reasoning workloads.
These works focus on the architecture and its full-precision kernels; the storage precision of the recurrent state is not their concern.

\paragraph{Quantization for ML models.}
Quantization has become a standard approach for reducing the memory requirements for LLM inference.
Weight-only methods such as GPTQ~\citep{frantar2022gptq} and AWQ~\citep{lin2024awq} compress the weights to 4 bits, and SmoothQuant~\citep{xiao2023smoothquant} further quantizes the activations by migrating their outliers into the weights.
As the context grows, the KV cache becomes the dominant memory cost. KIVI~\citep{liu2024kivi} quantizes keys per channel and values per token, KVQuant~\citep{hooper2024kvquant} isolates outliers into a sparse full-precision component.
SVDQuant~\citep{li2024svdquant} and GEAR~\citep{kang2024gear} absorb outliers into a high-precision low-rank component and quantize the residual.
For state space models, Quamba~\citep{chiang2025quamba} and MambaQuant~\citep{yue2025mambaquant} quantize the weights and activations, and Quamba2~\citep{chiang2025quamba2} and Q-Mamba~\citep{tianqi2025q} further quantize the cached SSM state.
These methods are specialized to Mamba's selective SSM, whose state is updated differently from that of other linear attention architectures.

\paragraph{State management in linear attention serving.}
Since the recurrent state is overwritten at every token, serving systems need extra mechanisms to keep or restore earlier states.
ReplaySSM~\citep{dao2026replayssm} supports speculative decoding by keeping the state at a checkpoint and recomputing the recent tokens for efficient rollback of the state.
Per-window quantization similarly starts from a stored boundary state and replays recent updates, but uses this structure to improve quantization accuracy.
Marconi~\citep{pan2025marconi} enables prefix caching for hybrid models by saving the states of cached prefixes as checkpoints, at the cost of a larger memory footprint.

\section{Method}
\label{sec:method}

\vspace{-0.5em}
\subsection{Preliminaries}
\label{sec:prelim}

We consider one head of a linear attention model with a recurrent state matrix $S_t\in\mathbb{R}^{d_k\times d_v}$, query $q_t\in\mathbb{R}^{d_k}$, and value and output $v_t,o_t\in\mathbb{R}^{d_v}$.
We study recurrences whose update is a diagonal decay followed by a single rank-one delta-rule~\citep{schlag2021linear} update:
\begin{equation}
  S_t = \Diag(\alpha_t)S_{t-1}
      + \beta_tk_t\bigl(v_t-S_{t-1}^{\top}(\alpha_t\odot k_t)\bigr)^{\top},
  \qquad o_t=S_t^{\top}q_t.
  \label{eq:general}
\end{equation}
Here $k_t$ is the key, $\beta_t$ the scalar write strength, and $\Diag(\alpha_t)$ the diagonal decay of KDA~\citep{team2025kimi}, whose erase $(I-\beta_tk_tk_t^{\top})$ is expanded; a scalar $\alpha_t$ gives Gated DeltaNet (Appendix~\ref{app:instances}).

For state quantization, denote $\hat S=\quant_b(S)$, where $\quant_b$ is the $b$-bit quantization operator, $S$ is the original state, and $\hat S$ is its quantized representation, including scale metadata.
We use $S_t^{\mathrm{FP32}}$ for the independent trajectory of \eqref{eq:general} without state quantization and $E_t=\dequant_b(\hat S_t)-S_t^{\mathrm{FP32}}$ for its quantization-induced deviation, where $\dequant_b$ is the corresponding dequantization operator.
When a quantized state participates in arithmetic below, its dequantization is implicit.

\subsection{Per-window quantization}
\label{sec:round1}

A na\"ive state-quantized model quantizes and dequantizes the full state after \emph{every} decode step.
At step $t$, it first computes the full-precision updated recurrent state \hlinline{fpmark}{S_t} from the previously quantized state \hlinline{quantmark}{\hat S_{t-1}}, then quantizes the resulting new recurrent state:
\begin{equation}
  \begin{aligned}
    \hfp{S_t} &= \Diag(\alpha_t)\,\hq{\hat S_{t-1}}
       + \beta_tk_t\bigl(v_t-\hat S_{t-1}^{\top}(\alpha_t\odot k_t)\bigr)^{\top},\\
    \hq{\hat S_t} &= \quant_b\bigl(\hfp{S_t}\bigr).
  \end{aligned}
  \label{eq:naive}
\end{equation}
\begin{wrapfigure}[15]{r}{0.40\linewidth}
    \vspace{-1\baselineskip}
    \centering
    \includegraphics[width=0.95\linewidth]{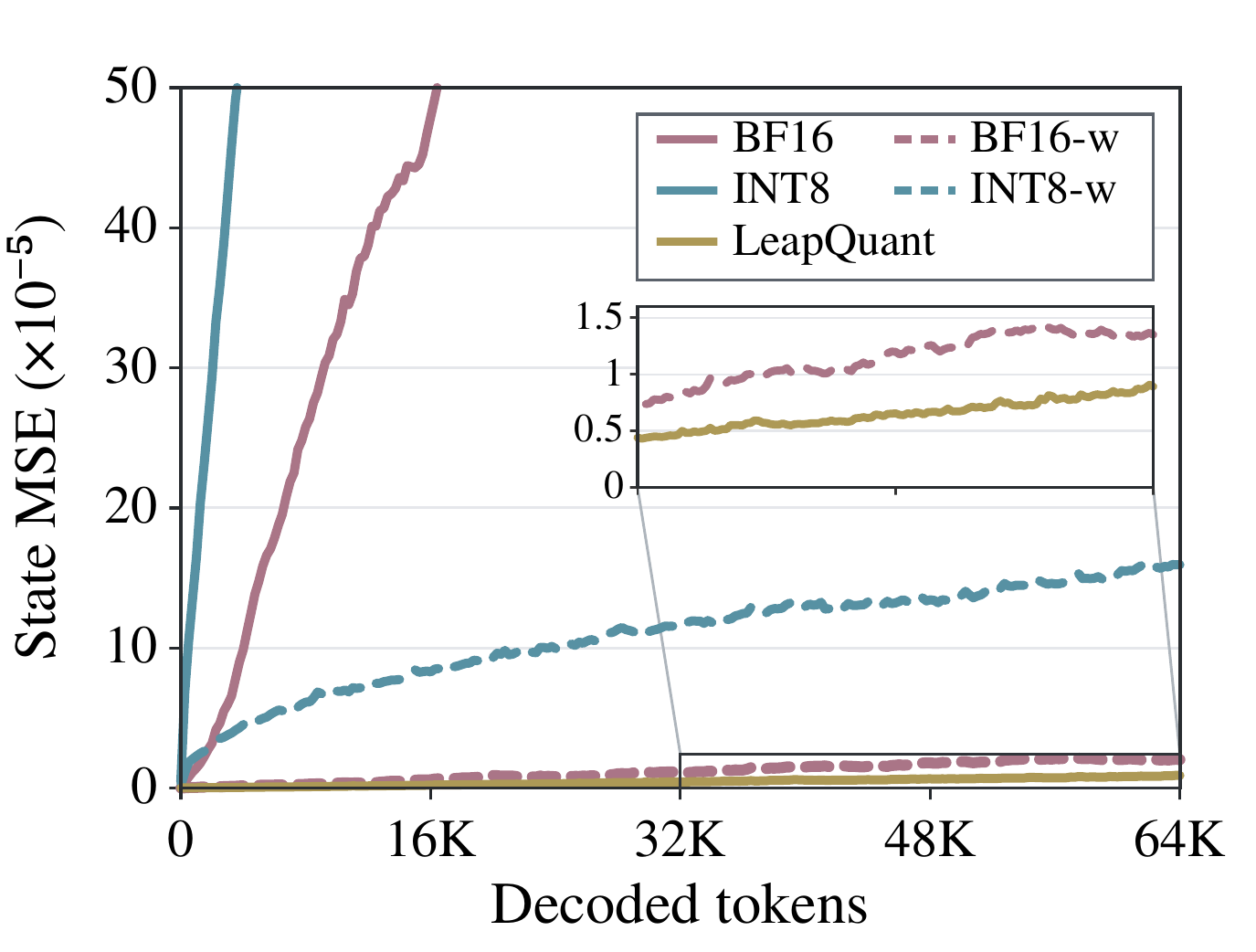}
    \caption{State MSE over 64K decoded tokens on PG-19 (Qwen3.5-9B); -w quantizes once per window.}
    \label{fig:state-error}
\end{wrapfigure}
\noindent Because every update starts from an already quantized state, its rounding error is carried into subsequent steps and can accumulate over long contexts (Appendix~\ref{app:deviation}).
Per-step quantization therefore introduces a new error at every token.
Figure~\ref{fig:state-error} tracks the error of the quantized state against the FP32 trajectory of Qwen3.5-9B over 64K decoded tokens.
Under per-step quantization, the error grows steadily with the context length, even for BF16.
Stochastic rounding helps little, as it only lowers the BF16 error at 64K by 3× and makes INT8 diverge.
In contrast, quantizing only once per 16-token window, as introduced below (the -w variants), lowers the error at 64K by $100\times$ for BF16 and by $39\times$ for INT8.
\method{} further reduces the error of each window-boundary quantization and keeps the lowest error throughout, even below per-window BF16 at half the bits.

\vspace{0.5\baselineskip}
Motivated by the error growth in Figure~\ref{fig:state-error}, we propose \textbf{per-window quantization} (Figure~\ref{fig:per-window}).
We write each update as a decay followed by a rank-one update whose correction vector $u_i$ is computed from the previous state.
For \eqref{eq:general} and the other models in Appendix~\ref{app:instances}, this update is described by the decay $\alpha_i$, the key $k_i$, and the computed correction $u_i$:
\begin{equation}
  u_i=\beta_i\bigl(v_i-S_{i-1}^{\top}(\alpha_i\odot k_i)\bigr),
  \qquad S_i=\Diag(\alpha_i)S_{i-1}+k_i u_i^{\top}.
  \label{eq:record}
\end{equation}
We now consider a sequence of $p$ consecutive tokens fed into the linear state.
For simplicity, index them as tokens $1,\ldots,p$.
We keep the quantized boundary state $\hat S_0$ fixed and buffer the $p$ tuples $(\alpha_i,k_i,u_i)$ in higher precision.
For $a\le b$, denote the cumulative decay from step $a$ through step $b$ by $\Gamma_{a:b}=\Diag(\alpha_b)\Diag(\alpha_{b-1})\cdots\Diag(\alpha_a)$, and set $\Gamma_{a:b}=I$ when $a>b$.
Then, for any $1\le\ell\le p$, unrolling \eqref{eq:record} from $\hat S_0$ yields (Appendix~\ref{app:readout})
\begin{equation}
  \begin{aligned}
    \hfp{S_\ell} &= \Gamma_{1:\ell}\,\hq{\hat S_0}
       + \sum_{j=1}^{\ell}\Gamma_{j+1:\ell}\,k_j u_j^{\top},
       \qquad 1\le\ell\le p,\\
    \hq{\hat S_p} &= \quant_b\bigl(\hfp{S_p}\bigr).
  \end{aligned}
  \label{eq:window}
\end{equation}
This is the chunk-level form that linear attention uses in training and prefill~\citep{yang2024parallelizing,yang2025gated}: intermediate states are expressed through the boundary state and the buffered tuples, so the full state is never re-quantized within the window.
Only at $\ell=p$ do we materialize $S_p$ and quantize it for the next window as $\hat S_p$.
This reduces the quantization frequency by a factor of $p$ and substantially lowers the state error over long contexts.

\begin{figure}[t]
    \centering
    \includegraphics[width=0.9\linewidth]{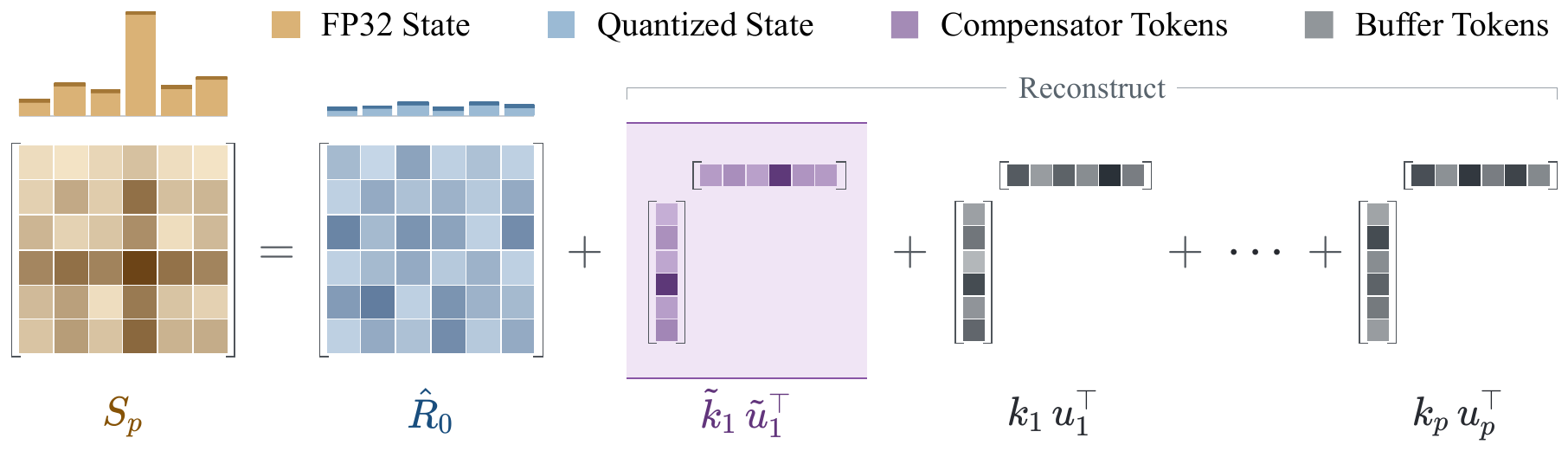}
    \caption{Per-window state reconstruction with Compensator Tokens. The FP32 state is split into a low-bit quantized residual, a few higher-precision Compensator Tokens that capture its dominant large-magnitude structure, and the buffered real-token updates of the window. The Compensator Tokens absorb dominant outliers, flattening the residual's row $\ell_2$ norms and making it easier to quantize.}
    \vspace{-1.5em}
    \label{fig:ghost-tokens}
\end{figure}

\subsection{Compensator Tokens}
\label{sec:round2}

\begin{figure}[t]
    \centering
    \begin{minipage}[t]{0.217\linewidth}
        \centering
        \vspace{0pt}
        \vspace{0.08\linewidth}
        \includegraphics[width=\linewidth]{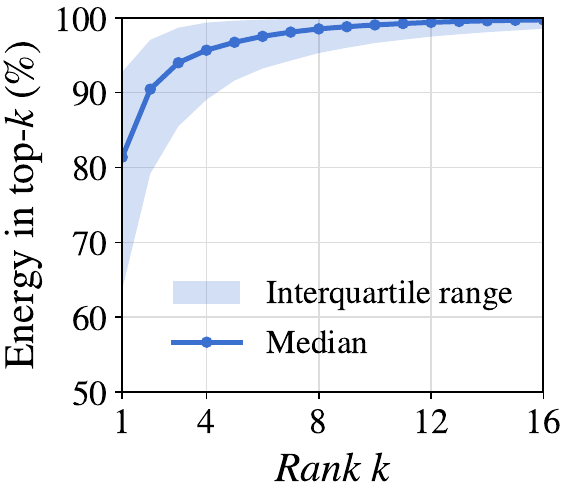}\\[0.17\linewidth]
        \caption{\protect\raggedright Energy of the top singular values in the Qwen3.5-9B model.}
        \vspace{-1.5em}
        \label{fig:energy}
    \end{minipage}%
    \hfill
    \begin{minipage}[t]{0.753\linewidth}
        \centering
        \vspace{0pt}
        \includegraphics[width=\linewidth]{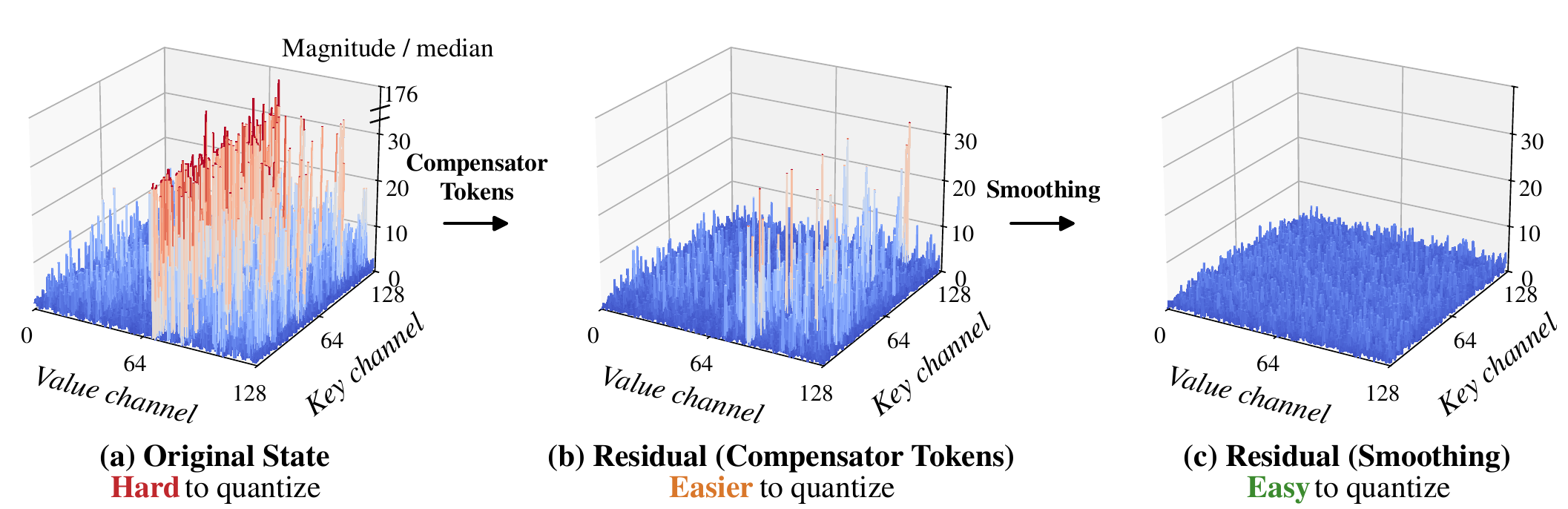}
        \caption{Magnitude of one Qwen3.5-9B head (layer 13, head 16) at 8K context, normalized by the median magnitude of $S$ in (a) and (b) and of the smoothed residual in (c).
        Outliers up to 176$\times$ the median (a) drop to 47$\times$ after Compensator Tokens (b) and to 5.3$\times$ after smoothing (c).}
        \vspace{-1.5em}
        \label{fig:state-heat}
    \end{minipage}
\end{figure}

Although per-window quantization reduces the quantization frequency, the reconstructed state $S_p$ can contain large-magnitude outliers concentrated in a few rows and columns (Figure~\ref{fig:state-heat}(a)), and a few singular components carry most of its energy (Figure~\ref{fig:energy}).
These outliers dominate the quantization scale, forcing most entries to be represented with unnecessarily coarse quantization steps.
To address this problem, we introduce \textbf{Compensator Tokens} (Figure~\ref{fig:ghost-tokens}), which reduce outlier-induced quantization error while integrating naturally with per-window quantization.

Before quantizing $S_p$, we fit a rank-one matrix $\tilde k\tilde u^{\top}$ to capture its dominant large-magnitude structure.
We subtract this matrix from $S_p$ and quantize only the residual $R_p$.
The largest values are therefore kept out of the tensor being quantized, reducing its dynamic range and alleviating the quantization pressure:
\begin{equation}
  \begin{aligned}
    (\tilde k,\tilde u)
       &= \operatorname*{arg\,min}_{k,u}
          \bigl\lVert\hfp{S_p}-ku^{\top}\bigr\rVert_F,\\
    R_p &= \hfp{S_p}-\tilde k\tilde u^{\top},
       \qquad \hq{\hat R_p}=\quant_b(R_p),\\
    \hrec{\tilde S_p} &= \dequant_b\bigl(\hq{\hat R_p}\bigr)
       +\tilde k\tilde u^{\top}.
  \end{aligned}
  \label{eq:compensator}
\end{equation}
Here $\lVert\cdot\rVert_F$ is the Frobenius norm.
We call the pair $(\tilde k,\tilde u)$ a \textbf{Compensator Token} as it preserves the dominant state component in higher precision and, together with identity decay, has the same rank-one update form as a real token in \eqref{eq:record}.
Unlike a real input token, it is not produced by the model and does not generate an output.
It exists only in the window representation and is placed before the real-token updates of the next window.
Since the decode kernel operates directly on rank-one updates, it processes the Compensator Token through the same path as the real tokens without any model-specific modification.

At the start of the next window, the quantized residual and the compensator tokens together represent the initial state.
The decode kernel processes the compensator tokens before the real-token updates, so subsequent updates and outputs use the combined state without the need for an extra kernel.

At the end of each window, we reconstruct $S_p$ from the quantized residual, the current compensator tokens, and the window's real-token updates.
We then discard the old compensator tokens, fit a new approximation to $S_p$, and quantize the remaining residual.
The new quantized residual and compensator tokens initialize the next window.

The construction extends directly to $r$ compensator tokens.
Let $\tilde K=[\tilde k_1,\ldots,\tilde k_r]\in\mathbb{R}^{d_k\times r}$ and $\tilde U=[\tilde u_1,\ldots,\tilde u_r]\in\mathbb{R}^{d_v\times r}$ collect their vector pairs.
Their combined contribution is
\begin{equation}
  \begin{aligned}
    \tilde K\tilde U^{\top}
       &= \sum_{h=1}^{r}\tilde k_h\tilde u_h^{\top},\\
    R_p &= \hfp{S_p}-\tilde K\tilde U^{\top},
       \qquad \hq{\hat R_p}=\quant_b(R_p),\\
    \hrec{\tilde S_p} &= \dequant_b\bigl(\hq{\hat R_p}\bigr)
       +\tilde K\tilde U^{\top}.
  \end{aligned}
  \label{eq:compensator-rank-r}
\end{equation}
In this way, compensator tokens preserve the dominant state structure in higher precision and leave a residual that is easier to quantize.
For the small values of $r$ used in practice, reconstructing the compensator tokens is fully hidden by the state memory read.
At each window boundary, we refit the compensator tokens on tensor cores without materializing the old ones in the dense state, so for $r\le 8$ the exposed kernel-level overhead stays within 7\% in all settings.

\subsection{Residual smoothing}
\label{sec:round3}

Compensator tokens preserve the dominant low-rank structure in higher precision, but the remaining residual can still have uneven magnitudes across key rows (Figure~\ref{fig:state-heat}(b)).
A few large key- and value-channels can dominate the quantization scale, leaving smaller entries with coarse resolution.
We therefore smooth the residual before quantization to balance its row magnitudes and make it easier to represent at low precision (Figure~\ref{fig:state-heat}(c)).

Let $R_0$ denote the residual at the start of a window.
Each entry of the smoothing vector $c\in\mathbb{R}_{>0}^{d_k}$ is the square root of the corresponding key row's mean absolute value, with a small positive floor to avoid division by zero.
Setting $C=\Diag(c)$, we left-multiply by $C^{-1}$ to balance the key-row magnitudes before quantization.
We quantize this rescaled residual and retain the smoothing scales alongside the quantized residual.
Since the distribution changes over time, these scales remain fixed within each window but are recomputed from the new residual at every boundary.

After dequantization, multiplying by $C$ restores the original key coordinates.
The transform itself is invertible; only the intervening quantization introduces approximation.
Adding back the unchanged higher-precision compensator tokens recovers the initial state.
Substituting this representation into \eqref{eq:window} gives the reconstructed state $S_p$ at the end of the window (Appendix~\ref{app:smoothing} derives the corresponding readout), without changing the buffered real-token updates:
\begin{equation}
  \begin{aligned}
    \hq{\hat R_0^C} &= \quant_b(C^{-1}R_0),\\
    \hfp{S_p} &= \Gamma_{1:p}
      \bigl[C\,\dequant_b\bigl(\hq{\hat R_0^C}\bigr)
          +\tilde K\tilde U^{\top}\bigr]
      +\sum_{j=1}^{p}\Gamma_{j+1:p}\,k_j u_j^{\top}.
  \end{aligned}
  \label{eq:smooth}
\end{equation}

\begin{table}[t]
  \centering
  \setlength{\tabcolsep}{2.4pt}
  \caption{Downstream performance (\%, higher is better), averaged over three seeds. Best scores in each column within the 8-bit, 6-bit, and 4-bit groups are in bold.}
  \renewcommand{\arraystretch}{1.1}
  \newcommand{\bmk}[1]{\multicolumn{1}{c}{\rotatebox[origin=lb]{45}{\scriptsize #1}}}
  \newcommand{\gp}{6pt}
  \newcommand{\bmkg}[1]{\multicolumn{1}{>{\hspace{\gp}}c}{\rotatebox[origin=lb]{45}{\scriptsize #1}}}
  \resizebox{\linewidth}{!}{%
  \begin{tabular}{@{}l>{\hspace{\gp}}r*{3}{r}>{\hspace{\gp}}r*{3}{r}>{\hspace{\gp}}r*{3}{r}>{\hspace{\gp}}r@{}}
    \toprule
    & \multicolumn{4}{>{\hspace{\gp}}c}{Qwen3.5-9B} & \multicolumn{4}{>{\hspace{\gp}}c}{GLM-5.3-Flash} & \multicolumn{4}{>{\hspace{\gp}}c}{Qwen3.8-Flash-Next} & \\
    \cmidrule(l{\gp}r){2-5}\cmidrule(l{\gp}r){6-9}\cmidrule(l{\gp}r){10-13}
    Method & \bmkg{AIME} & \bmk{Beyond} & \bmk{GPQA} & \bmk{LCB} & \bmkg{AIME} & \bmk{Beyond} & \bmk{GPQA} & \bmk{LCB} & \bmkg{AIME} & \bmk{Beyond} & \bmk{GPQA} & \bmk{LCB} & \multicolumn{1}{>{\hspace{\gp}}r}{Avg.} \\
    \midrule
    FP32 & 87.9 & 63.3 & 81.3 & 64.1 & 92.2 & 80.7 & 92.1 & 85.2 & 98.5 & 85.1 & 91.4 & 86.8 & 84.0 \\
    \midrule
    BF16 & 72.1 & 33.7 & 66.2 & 49.6 & 76.7 & 57.2 & 90.6 & 74.6 & 96.8 & 78.8 & 90.2 & 84.0 & 72.5 \\
    \midrule
    \multicolumn{14}{@{}l}{\textit{8-bit methods}} \\
    \rowcolor{oursrow}\textbf{Ours} & \textbf{87.9} & \textbf{64.0} & \textbf{81.8} & \textbf{64.1} & \textbf{91.1} & \textbf{83.2} & \textbf{91.2} & \textbf{87.5} & \textbf{99.0} & \textbf{85.3} & \textbf{91.6} & \textbf{85.8} & \textbf{84.4} \\
    KVQuant & 70.3 & 39.9 & 66.8 & 55.5 & 87.9 & 72.8 & 90.6 & 83.0 & 94.7 & 74.2 & 87.2 & 79.9 & 75.2 \\
    TurboQuant & 67.9 & 43.2 & 62.0 & 42.2 & 83.8 & 67.3 & 89.9 & 73.8 & 95.4 & 78.4 & 89.9 & 84.7 & 73.2 \\
    QuaRot & 11.4 & 7.4 & 38.2 & 35.6 & 91.0 & 76.2 & 90.9 & 78.6 & 88.3 & 61.0 & 84.7 & 72.0 & 61.3 \\
    INT8 & 3.9 & 1.8 & 17.2 & 10.7 & 87.4 & 66.8 & 90.4 & 67.4 & 67.5 & 41.8 & 61.3 & 56.0 & 47.7 \\
    FP8 & 0.0 & 0.1 & 2.2 & 5.9 & 69.6 & 44.2 & 82.0 & 53.2 & 82.5 & 52.8 & 76.8 & 71.5 & 45.1 \\
    \midrule
    \multicolumn{14}{@{}l}{\textit{6-bit methods}} \\
    \rowcolor{oursrow}\textbf{Ours} & \textbf{87.9} & \textbf{63.6} & \textbf{81.0} & \textbf{65.4} & \textbf{92.9} & \textbf{82.7} & \textbf{91.2} & \textbf{86.3} & \textbf{98.8} & \textbf{85.8} & \textbf{89.9} & \textbf{86.8} & \textbf{84.4} \\
    TurboQuant & 19.3 & 7.2 & 33.2 & 13.0 & 88.1 & 66.8 & 88.0 & 66.2 & 94.4 & 73.1 & 88.2 & 83.7 & 60.1 \\
    MXFP6 & 0.0 & 0.0 & 0.8 & 0.5 & 71.4 & 44.8 & 82.7 & 53.4 & 59.2 & 34.6 & 63.3 & 58.0 & 39.1 \\
    NVFP6 & 0.0 & 0.1 & 1.5 & 7.6 & 39.2 & 17.1 & 67.8 & 36.9 & 80.4 & 52.9 & 73.7 & 67.4 & 37.1 \\
    \midrule
    \multicolumn{14}{@{}l}{\textit{4-bit methods}} \\
    \rowcolor{oursrow}\textbf{Ours} & \textbf{85.0} & \textbf{60.1} & \textbf{80.0} & \textbf{61.3} & \textbf{85.0} & \textbf{67.6} & \textbf{91.6} & \textbf{72.8} & \textbf{92.5} & \textbf{72.6} & \textbf{90.6} & \textbf{79.9} & \textbf{78.3} \\
    TurboQuant & 3.5 & 1.4 & 26.8 & 11.5 & 59.7 & 43.2 & 77.9 & 49.4 & 79.3 & 49.8 & 74.9 & 62.8 & 45.0 \\
    MXFP4 & 0.0 & 0.0 & 3.7 & 0.3 & 25.3 & 18.6 & 70.9 & 29.3 & 1.7 & 2.8 & 25.4 & 22.6 & 16.7 \\
    \bottomrule
  \end{tabular}%
  }
  \label{tab:downstream}
\end{table}

\section{Evaluation}
\label{sec:evaluation}

\vspace{-0.5em}
\subsection{Setups}
\label{sec:eval-setup}

\paragraph{Models.}
We evaluate \method{} on five hybrid linear attention LLMs: Qwen3.5-9B, Qwen3.5-35B-A3B, and Qwen3.8-Flash-Next use GDN~\citep{yang2025gated}; Kimi-Linear-48B-A3B-Instruct~\citep{team2025kimi} and GLM-5.3-Flash~\citep{zai2026glm53flash} use KDA.

\paragraph{Datasets and Evaluation.}
We use AIME 2026~\citep{maa2026aime}, BeyondAIME~\citep{bytedance2025beyondaime}, HMMT~\citep{balunovic2025matharena}, GPQA-Diamond~\citep{rein2023gpqa}, MMLU-Pro~\citep{wang2024mmlu}, GSM8K~\citep{cobbe2021gsm8k}, LiveCodeBench v6~\citep{jain2025livecodebench}, USACO~\citep{shi2024usaco}, and OJBench~\citep{wang2025ojbench}.
We report downstream accuracy averaged over three random seeds.
For each model, all methods use the same prompts and the sampling parameters and output limits from its official model card (Appendix~\ref{app:eval-details}).
For decode and end-to-end efficiency, we compare against the FP32-state implementation of vLLM with CUDA graphs enabled.

\paragraph{Implementation.}
We implement \method{} in vLLM~\citep{kwon2023efficient} and write the linear attention decode kernels in TileLang~\citep{wang2025tilelang}. 
All other components (e.g., MoE layers) stay in the baseline precision.
At each window boundary, the compensator tokens are fitted with power iteration.
For all models and tasks, we use a window of $p=16$ tokens, $r=4$ FP16 Compensator Tokens ($r=6$ at 6 and 4 bits), and FP32 smoothing scales (Section~\ref{sec:ablation} studies $p$ and $r$).
All experiments run on NVIDIA B200, RTX PRO 6000, and RTX 5090 GPUs.

\paragraph{Baselines.}
We compare \method{} with the FP32 and BF16 states, the default and optional formats in vLLM and SGLang, and with state quantization baselines that re-quantize the state after every decode step.
At 8 bits, we evaluate FP8 and INT8 at their best granularities.
Given the lack of existing recurrent state quantization methods, we also evaluate three methods originally designed for the KV cache or activations, which we adapt to the recurrent state: KVQuant~\citep{hooper2024kvquant}, QuaRot~\citep{ashkboos2024quarot}, and TurboQuant~\citep{zandieh2026turboquant}.
At 6 and 4 bits, we evaluate NVFP6, MXFP6, and INT6, and NVFP4, MXFP4, and per-channel INT4, together with low-bit variants of the three adapted methods.
Table~\ref{tab:downstream} reports the best direct and adapted methods at each of these bit widths, and Appendix~\ref{app:full-results} lists all of them.

\subsection{Accuracy Results}
\label{sec:downstream}
\paragraph{Downstream tasks.}
Table~\ref{tab:downstream} shows the downstream accuracy of the three models.
With 8-bit quantization, \method{} achieves accuracy on par with the FP32 baseline across the 12 model--task pairs.
It stays within 2.5\% of FP32 on every pair and averages 84.4\%, versus 84.0\% for FP32.
In contrast, per-step quantization leads to poor accuracy even at 16 bits: storing the Qwen3.5-9B state in BF16 lowers its AIME score from $87.9\%$ to $72.1\%$, whereas \method{} keeps $87.9\%$ with half as many bits.
This gap is especially clear on long generations: Qwen3.5-9B's AIME reasoning traces can run to tens of thousands of tokens, and per-step FP8 and INT8 score below 4\% on AIME.
GLM-5.3-Flash and Qwen3.8-Flash-Next are less sensitive to per-step errors, but every 8-bit baseline still loses more than 10\% on at least one of their tasks, and the best one, KVQuant, averages only 75.2\%.

\paragraph{Lower bit widths.}
The gap between per-window and per-step quantization widens as the bit width decreases.
At 6 bits, the best baseline, TurboQuant, averages only 60.1\% and collapses on Qwen3.5-9B, while \method{} still averages 84.4\%, on par with FP32 and above every 8-bit baseline.
At 4 bits, TurboQuant and MXFP4 average 45.0\% and 16.7\%, while \method{} averages 78.3\%.
It stays within 3.2\% of FP32 on every task of Qwen3.5-9B; GLM-5.3-Flash and Qwen3.8-Flash-Next are more sensitive at this bit width, with the largest drops on BeyondAIME and LiveCodeBench.
These results show that \method{} retains substantially more downstream accuracy than the baselines even under aggressive 4-bit state quantization.

\subsection{Efficiency Results}
\label{sec:efficiency}

\begin{figure}[t]
    \centering
    \includegraphics[width=\linewidth]{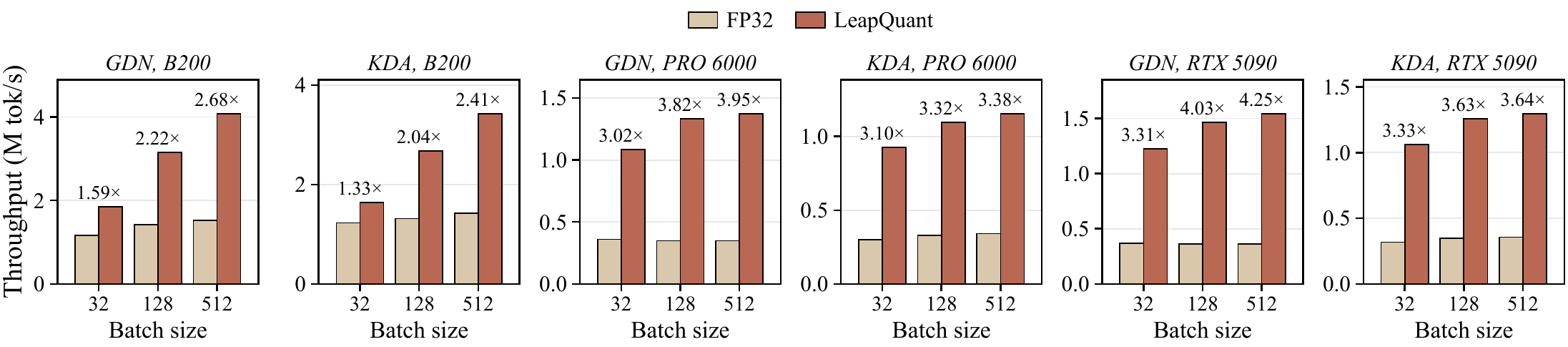}
    \caption{Kernel throughput of one linear attention layer.}
    \label{fig:eval-kernel}
\end{figure}

\begin{figure}[t]
    \centering
    \includegraphics[width=\linewidth]{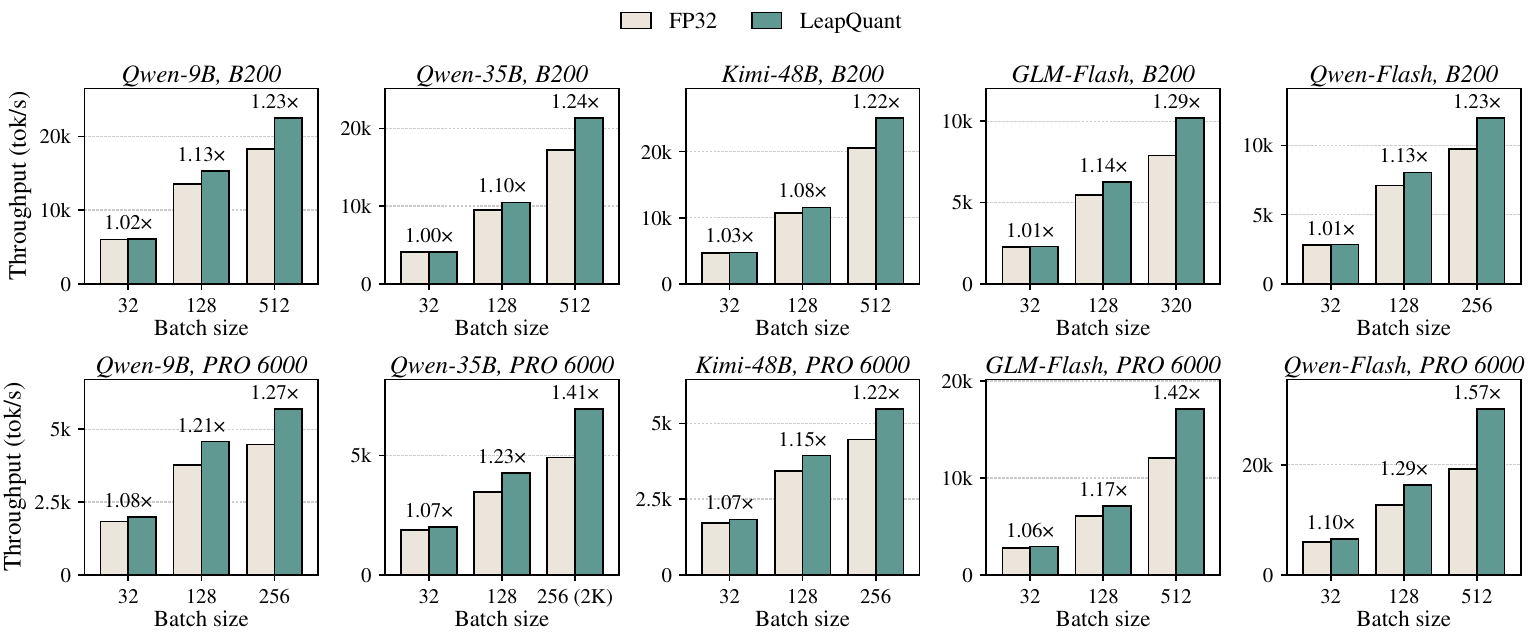}
    \caption{Decode-step throughput in vLLM at context length 4K. On the RTX PRO 6000 with limited memory capacity, we use a smaller maximum batch size and context length, and we only use 8 layers of GLM-5.3-Flash and Qwen3.8-Flash-Next.}
    \label{fig:eval-e2e}
\end{figure}

\paragraph{Kernel speedup.}
Figure~\ref{fig:eval-kernel} compares one linear attention layer (32 heads, $d_k=d_v=128$) with the FP32 kernel in FLA.
At batch size 512, \method{} is 2.68$\times$, 3.95$\times$, and 4.25$\times$ faster on GDN and 2.41$\times$, 3.38$\times$, and 3.64$\times$ on KDA on the B200, RTX PRO 6000, and RTX 5090.
The larger gains on the RTX PRO 6000 and RTX 5090 are consistent with recurrent-state traffic being a stronger bottleneck on these GPUs.

\paragraph{Decode speedup.}
Figure~\ref{fig:eval-e2e} measures pure decode steps on five hybrid models.
At the largest batch size per GPU on the B200, \method{} improves decode throughput by 1.22--1.29$\times$; at the largest batch size that fits on the RTX PRO 6000, the gain is 1.22--1.57$\times$. The gain increases with batch size as linear attention takes a larger share of each step.
Appendix~\ref{app:efficiency} reports other context lengths, up to 128K.

\paragraph{Memory reduction.}
\method{} stores each window-boundary state at 1.19 bytes per element instead of 4, including the INT8 residual, the smoothing scales, and the Compensator Tokens, a 3.4$\times$ reduction.
At a window length of 16, the BF16 update buffer requires only around 8 KiB per head per layer for 128$\times$128 states and is allocated once per active request, making its overhead marginal.
When prefix caching is enabled in the default mode in vLLM and SGLang, a checkpoint is kept per 1K cached tokens.
In this setting, \method{} reduces the end-to-end memory by 41\%, 51\%, and 56\% on Qwen3.5-9B, Qwen3.5-35B-A3B, and Kimi-Linear-48B-A3B.
This allows up to 1.4$\times$ more concurrent requests when serving the Qwen3.5-9B model on one B200 GPU.

\begin{figure}[t]
    \centering
    \includegraphics[width=\linewidth]{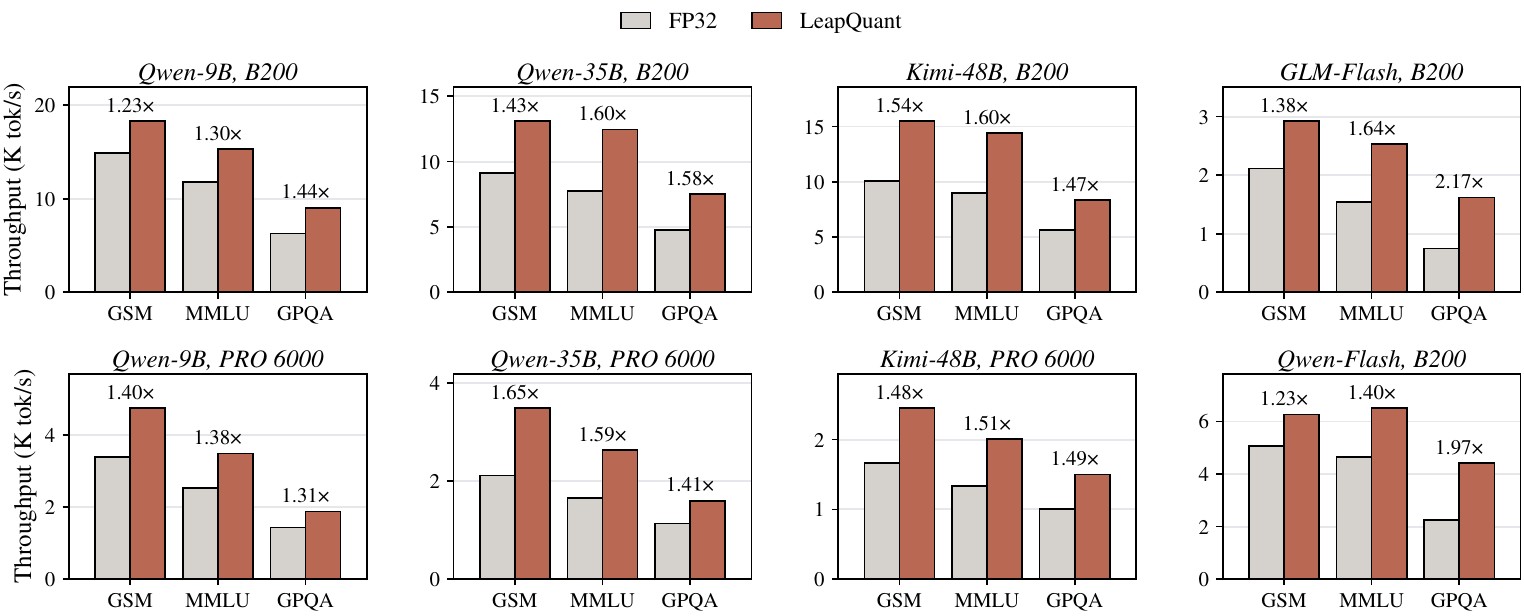}
    \caption{End-to-end inference throughput on different datasets. GLM-5.3-Flash and Qwen3.8-Flash-Next run on two GPUs with tensor parallelism, and their throughput is per GPU.}
    \vspace{-1em}
    \label{fig:prefix}
\end{figure}

\paragraph{End-to-end inference speedup.}
We run offline inference on GSM8K, MMLU-Pro, and GPQA with the input and output lengths of the original model.
Since Kimi-Linear is a non-reasoning model, we use the distribution of Qwen3.5-9B for it to simulate the real workload of production KDA models.
With more concurrent requests and faster decode steps, \method{} improves the output throughput by 1.23--2.17$\times$ on the B200 and 1.31--1.65$\times$ on the RTX PRO 6000 (Figure~\ref{fig:prefix}).
The gain is largest on GPQA, whose long outputs make the state checkpoints dominate memory: \method{} raises the average number of concurrent requests from 24 to 50 on GLM-5.3-Flash and from 63 to 115 on Qwen3.8-Flash-Next.

\vspace{-0.5em}
\subsection{Ablation Study}
\label{sec:ablation}

We ablate the design choices of \method{} on Qwen3.5-9B, using AIME 2026 and LiveCodeBench v6 as the most sensitive tasks in Table~\ref{tab:downstream}; kernel speedups are over the FP32 kernel at batch size 256.

\begin{wraptable}[13]{r}{0.43\linewidth}
  \vspace{-13pt}
  \raggedright
  \caption{\protect\raggedright Ablation of individual \method{} components on Qwen3.5-9B.}
  \label{tab:ablation}
  \centering
  \small
  \setlength{\tabcolsep}{0.5pt}
  \begin{tabular*}{\linewidth}{@{\extracolsep{\fill}}lccc@{}}
    \toprule
    Method & AIME & LCB & Kernel\\
    \midrule
    FP32 per-step & \textbf{87.9} & \textbf{64.1} & 1.00$\times$\\
    \midrule
    BF16 per-step & 72.1 & 49.6 & 1.64$\times$\\
    + Per-window & \textbf{87.8} & 62.9 & 1.71$\times$\\
    \midrule
    INT8 per-step & 7.1 & 9.2 & 2.43$\times$\\
    + Per-window & 82.4 & 60.6 & 2.64$\times$\\
    + Comp.\ Tokens & 86.6 & 61.5 & 2.56$\times$\\
    + Smoothing & \textbf{87.9} & \textbf{64.1} & 2.52$\times$\\
    \bottomrule
  \end{tabular*}
  \vspace{-8pt}
\end{wraptable}
\paragraph{Effectiveness of each design.}
Table~\ref{tab:ablation} adds the designs one at a time to an INT8 state, with BF16 as a higher-precision reference.
Per-step INT8 drops AIME from 87.9\% to 7.1\% and LiveCodeBench from 64.1\% to 9.2\%.
Per-window quantization improves AIME and LiveCodeBench accuracy to 82.4\% and 60.6\%, respectively; with BF16, it also raises AIME from 72.1\% to 87.8\%.
Compensator Tokens then improve the INT8 scores to 86.6\% and 61.5\%, while smoothing closes the remaining gap to FP32 at 2.52$\times$ kernel speedup.
Each stage improves quality while the complete 8-bit method retains a substantial efficiency advantage.
For efficiency, we compare against the FP32 and BF16 per-step FLA kernels and the BF16 ReplaySSM kernel in the official repository~\citep{dao2026replayssm}.
With all three components, \method{} matches FP32 accuracy at a 2.52$\times$ kernel speedup.

\paragraph{Window length.}
Table~\ref{tab:ablation-pr} (left) varies the window size $p$ with $r=4$.
Longer windows quantize less often, so accuracy improves up to $p=16$ and then saturates.
The kernel is fastest at $p=16$: shorter windows reconstruct the state more often, while longer ones enlarge the record buffer read at every step and leave less shared memory for pipelining.
$p=32$ gives no further accuracy gain, but its kernel speedup drops from 2.52$\times$ to 2.20$\times$ due to the increased memory traffic.

\begin{wraptable}{r}{0.48\linewidth}
  \vspace{-2pt}
  \raggedright
  \caption{\protect\raggedright Ablations of window length $p$ and Compensator Token count $r$ on Qwen3.5-9B.}
  \label{tab:ablation-pr}
  \centering
  \begin{minipage}[t]{0.48\linewidth}
    \centering
    \small
    \setlength{\tabcolsep}{0.5pt}
    \begin{tabular*}{\linewidth}{@{\extracolsep{\fill}}cccc@{}}
      \toprule
      $p$ & AIME & LCB & Kernel \\
      \midrule
      4 & 86.2 & 62.8 & 1.59$\times$ \\
      8 & 87.1 & 63.2 & 2.13$\times$ \\
      16 & \textbf{87.9} & \textbf{64.1} & \textbf{2.52$\times$} \\
      32 & \textbf{87.9} & \textbf{64.2} & 2.20$\times$ \\
      \bottomrule
    \end{tabular*}
  \end{minipage}\hfill%
  \begin{minipage}[t]{0.48\linewidth}
    \centering
    \small
    \setlength{\tabcolsep}{0.5pt}
    \begin{tabular*}{\linewidth}{@{\extracolsep{\fill}}cccc@{}}
      \toprule
      $r$ & AIME & LCB & Kernel \\
      \midrule
      2 & 85.7 & 63.4 & \textbf{2.57$\times$} \\
      4 & 87.9 & \textbf{64.1} & 2.52$\times$ \\
      8 & \textbf{88.1} & \textbf{64.0} & 2.37$\times$ \\
      16 & 87.9 & 62.8 & 0.74$\times$ \\
      \bottomrule
    \end{tabular*}
  \end{minipage}
  \vspace{-8pt}
\end{wraptable}
\paragraph{Number of Compensator Tokens.}
Table~\ref{tab:ablation-pr} (right) varies $r$ with $p=16$.
Accuracy saturates at $r=4$, consistent with the energy concentrated in a few singular values (Figure~\ref{fig:energy}).
Efficiency-wise, the power iteration and reconstruction work of up to four compensator tokens is fully overlapped with the memory reads on the B200 GPU, but it becomes exposed for larger $r$ and makes the kernel even slower than FP32 at $r=16$.
We therefore use $r=4$, which matches the accuracy of larger $r$ at nearly the kernel speed of $r=2$.

\vspace{-0.75em}
\section{Conclusion}
\label{sec:conclusion}

We present \method{}, a training-free method for near-lossless quantization of recurrent states in linear attention.
\sys{} mitigates error accumulation by quantizing the state only once per window, and captures state outliers in a few high-precision compensator tokens before smoothing the residual.
With an 8-bit state, \method{} matches FP32 accuracy on long reasoning and code generation while reducing the state memory and accelerating decoding on both GDN and KDA models.
As hybrid models devote more of their layers to linear
attention, we believe \method{} can make low-precision recurrent states a practical default for serving them.
\section*{Acknowledgement}

We thank Shu Liu, Hanfei Yu, Yichuan Wang, and Melissa Pan, as well as other members of the UC Berkeley Sky Computing Lab, for their insightful discussions during the early stages of this work.
This research is supported by NSF (IFML) CCF-2019844 and gifts from Accenture, AMD, Anyscale, Broadcom Inc., Google, IBM, Intel, Intesa Sanpaolo, Lambda, Mibura Inc., Samsung SDS, and SAP.

\bibliography{bib/reference}
\bibliographystyle{iclr2027_conference}

\newpage
\appendix
\section{Instances of the General Update}
\label{app:instances}

To cover more linear attention models, we write the update in a general form with the diagonal decay $\Diag(\alpha_t)$, a write vector $w_t$, and a read vector $b_t$:
\begin{equation}
  S_t = \Diag(\alpha_t) S_{t-1} + w_t\bigl(v_t - S_{t-1}^{\top} b_t\bigr)^{\top},
  \qquad
  o_t = S_t^{\top} q_t.
  \label{eq:general-app}
\end{equation}
The main text's update \eqref{eq:general} is the case $w_t=\beta_tk_t$ and $b_t=\alpha_t\odot k_t$.
Table~\ref{tab:instances} lists common instances: linear attention~\citep{katharopoulos2020transformers}, RetNet~\citep{sun2023retentive}, Mamba2~\citep{dao2024transformers}, GLA~\citep{yang2023gated}, RWKV6~\citep{peng2024eagle}, HGRN2~\citep{qin2024hgrn2}, DeltaNet~\citep{yang2024parallelizing}, Gated DeltaNet~\citep{yang2025gated}, KDA~\citep{team2025kimi} and DeltaProduct~\citep{siems2025deltaproduct}.
The decay may be the identity, a constant scalar, a data-dependent scalar or a data-dependent vector.
The term $S_{t-1}^{\top}b_t$ reads the previous state, and the write subtracts it from $v_t$ along the same direction $w_t$; this is the delta rule.
Plain and gated linear attention have $b_t=0$.

\begin{table}[h]
  \centering
  \caption{Instances of \eqref{eq:general-app}. DeltaProduct applies $n$ delta-rule sub-steps per token, each of which is an instance of \eqref{eq:general-app}. The last column gives the number of values in one buffered update $(\alpha_i,w_i,u_i)$.}
  \label{tab:instances}
  \small
  \begin{tabular}{@{}lcccc@{}}
    \toprule
    Model & Decay & $w_t$ & $b_t$ & Record size \\
    \midrule
    Linear Attention & $I$ & $k_t$ & $0$ & $d_k+d_v$ \\
    RetNet & $\gamma I$ & $k_t$ & $0$ & $d_k+d_v$ \\
    Mamba2 & $\alpha_t I$ & $k_t$ & $0$ & $1+d_k+d_v$ \\
    GLA / RWKV6 / HGRN2 & $\Diag(\alpha_t)$ & $k_t$ & $0$ & $2d_k+d_v$ \\
    DeltaNet & $I$ & $\beta_t k_t$ & $k_t$ & $d_k+d_v$ \\
    Gated DeltaNet & $\alpha_t I$ & $\beta_t k_t$ & $\alpha_t k_t$ & $1+d_k+d_v$ \\
    KDA & $\Diag(\alpha_t)$ & $\beta_t k_t$ & $\alpha_t\odot k_t$ & $2d_k+d_v$ \\
    DeltaProduct, sub-step $j$ & $I$ & $\beta_{t,j}k_{t,j}$ & $k_{t,j}$ & $d_k+d_v$ \\
    \bottomrule
  \end{tabular}
\end{table}

\paragraph{Gated DeltaNet and KDA.}
Expanding the original updates recovers the rows of Table~\ref{tab:instances}:
\begin{align*}
  \text{GDN:}\quad
  \alpha_t\bigl(I-\beta_tk_tk_t^{\top}\bigr)S_{t-1}+\beta_tk_tv_t^{\top}
  &= \alpha_tS_{t-1} + \beta_tk_t\bigl(v_t - S_{t-1}^{\top}(\alpha_tk_t)\bigr)^{\top},\\
  \text{KDA:}\quad
  \bigl(I-\beta_tk_tk_t^{\top}\bigr)\Diag(\alpha_t)S_{t-1}+\beta_tk_tv_t^{\top}
  &= \Diag(\alpha_t)S_{t-1} + \beta_tk_t\bigl(v_t - S_{t-1}^{\top}(\alpha_t\odot k_t)\bigr)^{\top}.
\end{align*}
In both cases the erase direction coincides with the write direction, so each token contributes a single rank-one record.
Here the write strength $\beta_t$ is carried by $w_t$; the main text instead places it in the correction, writing along $k_t$ with $u_t=\beta_t(v_t-S_{t-1}^{\top}b_t)$, which is equivalent.

\paragraph{Models with a separate erase direction.}
The form \eqref{eq:general-app} requires the erase direction to coincide with the write direction.
A model whose erase direction differs from its write direction still fits by splitting each token into two sub-steps of \eqref{eq:general-app}.
For RWKV7~\citep{peng2025rwkv}, whose transition is $\Diag(\alpha_t)-\hat{\kappa}_t(a_t\odot\hat{\kappa}_t)^{\top}$ with decay $\alpha_t$, the two sub-steps are an erase with $v=0$ and a pure write with identity decay and $b=0$:
\begin{align*}
  S_t' &= \Diag(\alpha_t)\,S_{t-1} + \hat{\kappa}_t\bigl(0 - S_{t-1}^{\top}(a_t\odot\hat{\kappa}_t)\bigr)^{\top},\\
  S_t &= S_t' + k_t v_t^{\top}.
\end{align*}
Each token then contributes two rank-one records, the same treatment as a DeltaProduct token with $n=2$.

\section{Derivations}
\label{app:derivations}

\subsection{Propagation of the quantization-induced deviation}
\label{app:deviation}

For this recurrence-level analysis, we fix the layer-input sequence and supply the FP32 reference and the quantized recurrence with the same coefficients $(\alpha_t,w_t,b_t,v_t)$ of \eqref{eq:general-app}:
\begin{align*}
  S_t^{\mathrm{FP32}} &= \Diag(\alpha_t)S_{t-1}^{\mathrm{FP32}} + w_t\bigl(v_t - (S_{t-1}^{\mathrm{FP32}})^{\top}b_t\bigr)^{\top},\\
  S_t &= \Diag(\alpha_t)\Sh_{t-1} + w_t\bigl(v_t - \Sh_{t-1}^{\top}b_t\bigr)^{\top}.
\end{align*}
Subtracting the two, the value $v_t$ cancels:
\begin{equation*}
  S_t - S_t^{\mathrm{FP32}} = \Diag(\alpha_t)E_{t-1} - w_t\bigl(E_{t-1}^{\top}b_t\bigr)^{\top}.
\end{equation*}
Adding $\varepsilon_t=\DQ_b(\Sh_t)-S_t$ to both sides gives
\begin{equation}
  E_t=\Diag(\alpha_t)E_{t-1}-w_t\bigl(E_{t-1}^{\top}b_t\bigr)^{\top}+\varepsilon_t.
  \label{eq:deviation}
\end{equation}
Unrolling it, each error $\varepsilon_s$ is carried to step $t$ by the same decay and erase steps that act on memories written at step $s$, and it reaches the output through $o_t=S_t^{\top}q_t$.
Under per-window quantization, $\varepsilon_s$ is nonzero only at window boundaries.
Because each update is small relative to the state, per-step rounding discards part of it at every step, so these errors are correlated and add up coherently.
Per-window quantization rounds the accumulated updates only once, which reduces the error by far more than the factor of $p$ in quantization frequency.

\subsection{Window unrolling and state readout}
\label{app:readout}

Restoring an arbitrary window-start index $t$, \eqref{eq:window} follows from \eqref{eq:record} by induction on the window length: applying one more record multiplies every existing term by $\Diag(\alpha_{t+p+1})$ and appends $k_{t+p+1}u_{t+p+1}^{\top}$, and $\Diag(\alpha_{t+p+1})\Gamma_{a:t+p}=\Gamma_{a:t+p+1}$.
Transposing the unrolled state at step $i$ and multiplying by $x$ gives, for $t\le i\le t+p$,
\begin{equation}
  S_i^{\top}x=\Sh_t^{\top}\Gamma_{t+1:i}x
    +\sum_{j=t+1}^{i}u_j\bigl(k_j^{\top}\Gamma_{j+1:i}x\bigr),
  \label{eq:readout}
\end{equation}
using $\Gamma^{\top}=\Gamma$ for diagonal $\Gamma$.
The first term costs $O(d_kd_v)$, and each of the $i-t$ records costs $O(d_k+d_v)$: one inner product $k_j^{\top}(\Gamma_{j+1:i}x)$ and one scaled vector $u_j$.
The diagonal products $\Gamma_{j+1:i}$ are cumulative decays and can be maintained incrementally, so the read never forms a $d_k\times d_v$ intermediate state.

\subsection{Readout with a smoothed quantized residual}
\label{app:smoothing}

For per-value-channel symmetric INT8 quantization, let $Z_0$ denote the integer payload and $B$ the diagonal scale matrix, so that $\dequant_b(\hat R_0^C)=(Z_0/127)B$.
Therefore, for any read vector $x$ and cumulative diagonal decay $\Gamma$,
\begin{equation*}
  \bigl[C\,\dequant_b(\hat R_0^C)\bigr]^{\top}\Gamma x
    =\left(C\frac{Z_0}{127}B\right)^{\top}\Gamma x
    =B\frac{Z_0^{\top}}{127}C\Gamma x.
\end{equation*}
$C$ and $\Gamma$ act on the key coordinates, while $B$ rescales the quantized residual's value-side projection.
Neither scale is applied to the real or Compensator Token records, whose contributions are already expressed in the original state coordinates.

For the Compensator Token component $\tilde K\tilde U^{\top}$, the corresponding read is
\begin{equation*}
  (\tilde K\tilde U^{\top})^{\top}\Gamma x
    =\tilde U(\tilde K^{\top}\Gamma x)
    =\sum_{h=1}^{r}\tilde u_h(\tilde k_h^{\top}\Gamma x).
\end{equation*}
Adding these rank-one records and the real records to the quantized-residual read gives the readout form of \eqref{eq:smooth}:
\begin{equation}
  \begin{aligned}
    S_\ell^{\top}x
      ={}&\dequant_b(\hat R_0^C)^{\top}C\Gamma_{1:\ell}x
        +\tilde U\bigl(\tilde K^{\top}\Gamma_{1:\ell}x\bigr)\\
       &+\sum_{j=1}^{\ell}u_j\bigl(k_j^{\top}\Gamma_{j+1:\ell}x\bigr).
  \end{aligned}
  \label{eq:smooth-readout}
\end{equation}
At a boundary, the sum is reconstructed in the original state coordinates before fitting the new factors and recomputing the residual scales.

\section{Additional Efficiency Evaluation}
\label{app:efficiency}

\begin{figure}[h]
    \centering
    \includegraphics[width=\linewidth]{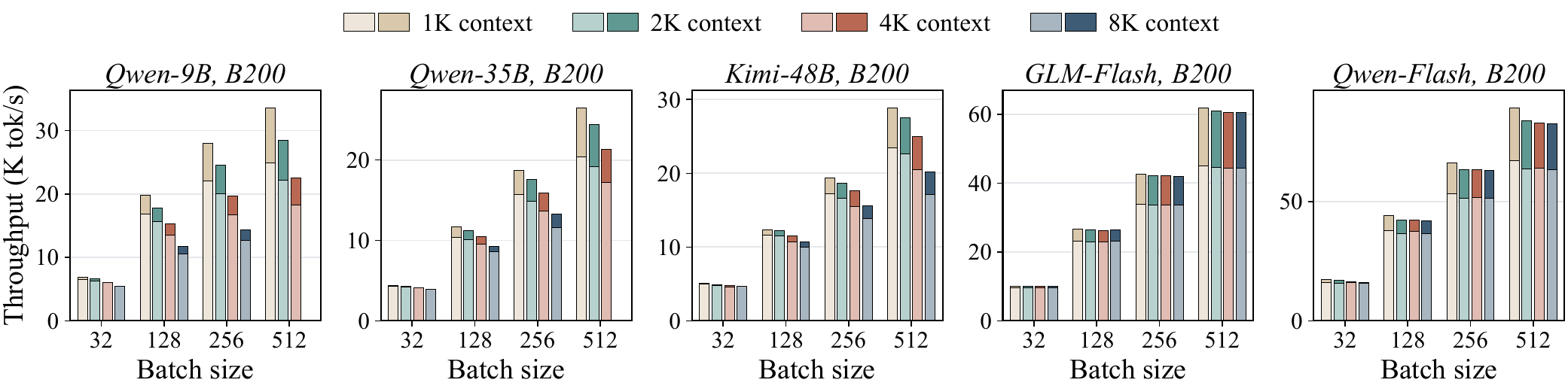}
    \caption{Decode-step throughput in vLLM on the B200 at context lengths 1K--8K. Each bar stacks the FP32 throughput (lighter) and the gain of \method{} (darker).}
    \vspace{-1em}
    \label{fig:app-bs}
\end{figure}

\begin{figure}[h]
    \centering
    \includegraphics[width=0.85\linewidth]{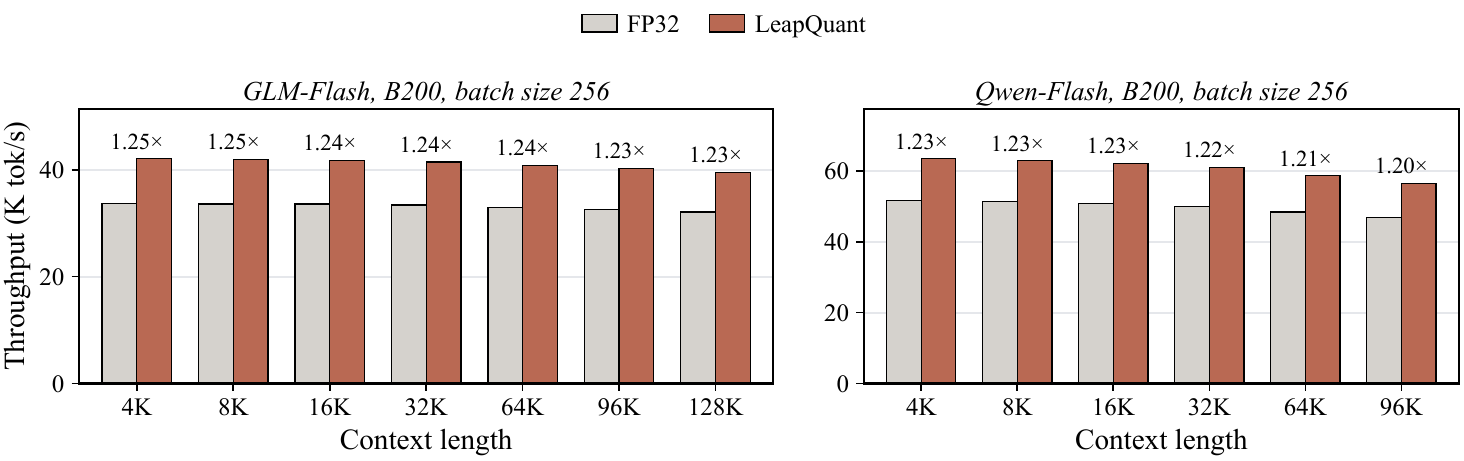}
    \caption{Decode-step throughput of the two sparse-attention models on the B200 at batch size 256 and long contexts.}
    \label{fig:app-long}
\end{figure}

Figure~\ref{fig:app-bs} reports the decode-step throughput on the B200 at context lengths from 1K to 8K, complementing the 4K results in Figure~\ref{fig:eval-e2e}.
Each bar stacks the FP32 throughput (lighter shade) and the gain of \method{} (darker shade).
For models that pair linear attention with dense full attention, the gain shrinks slightly as the context grows, since the full attention layers take a growing share of each step.
It remains substantial at 8K, where \method{} still accelerates Kimi-Linear-48B-A3B by 1.18$\times$ at batch size 512.
Recent models increasingly combine linear attention with sparse attention rather than full attention, which keeps the attention cost nearly constant as the context grows.
Figure~\ref{fig:app-long} evaluates two such models, 8-layer versions of GLM-5.3-Flash and Qwen3.8-Flash-Next, up to 128K and 96K at batch size 256.
\method{} accelerates them by 1.25$\times$ and 1.23$\times$ at 4K and still by 1.23$\times$ and 1.20$\times$ at the longest context.
We therefore expect \method{} to perform equally well at very long contexts on upcoming hybrid models.

\section{Evaluation Details}
\label{app:eval-details}

\paragraph{Sampling parameters.}
Table~\ref{tab:sampling} lists the generation settings.
All methods of a model share the same settings, prompts, and sample set.
We follow the benchmark settings of the official model cards: thinking is enabled for the Qwen3.5 models, GLM-5.3-Flash uses the maximum reasoning effort, and Qwen3.8-Flash-Next uses the extra-high reasoning effort; Kimi-Linear-48B-A3B-Instruct is an instruction model without a thinking mode.
The maximum model length is set above the output limit, so that no generation is truncated by the context window before reaching the output limit.
For Qwen3.5-9B, GLM-5.3-Flash, and Qwen3.8-Flash-Next, we stop a generation early once its last 32,768 tokens repeat verbatim with a period of at most 512 tokens and count it as truncated; on earlier runs without this rule, none of the outputs it would stop reached a correct answer.
We also cap the number of concurrent requests so that no request is preempted, since recomputing a preempted request restores its exact state and would favor the per-step baselines.

\begin{table}[h]
  \centering
  \small
  \caption{Generation settings of the downstream evaluation.}
  \label{tab:sampling}
  \begin{tabular}{@{}lcccccc@{}}
    \toprule
    Model & Temp. & Top-$p$ & Top-$k$ & Presence & Thinking & Max output \\
    \midrule
    Qwen3.5-9B & 1.0 & 0.95 & 20 & 1.5 & on & 81{,}920 \\
    GLM-5.3-Flash & 1.0 & 0.95 & -- & 0 & max & 327{,}680 \\
    Qwen3.8-Flash-Next & 1.0 & 0.95 & 20 & 0 & xhigh & 131{,}072 \\
    Qwen3.5-35B-A3B & 1.0 & 0.95 & 20 & 1.5 & on & 81{,}920 \\
    Kimi-Linear-48B-A3B & 1.0 & 1.0 & -- & 0 & -- & 65{,}536 \\
    \bottomrule
  \end{tabular}
\end{table}

\section{Full Downstream Results}
\label{app:full-results}

Table~\ref{tab:full-results} extends Table~\ref{tab:downstream} with HMMT, MMLU-Pro, USACO, and OJBench, and with every direct quantization format and adapted method we evaluate at 8, 6, and 4 bits.
Direct quantization formats are listed before the adapted methods in each group, and all baselines re-quantize the state at every decode step.
\method{} uses the configurations of Section~\ref{sec:eval-setup} at each bit width.
USACO and OJBench are single runs; all other scores are averaged over three seeds.
The table also lists the 8-bit results of Qwen3.5-35B-A3B and Kimi-Linear-48B-A3B.

\makeatletter
\ifdefined\apM\else\newdimen\apM\fi
\ifdefined\apW\else\newdimen\apW\fi
\ifdefined\apRW\else\newdimen\apRW\fi
\ifdefined\apRM\else\newdimen\apRM\fi
\ifdefined\apLH\else\newdimen\apLH\fi
\ifdefined\apS\else\newdimen\apS\fi
\begin{table}[p]
  \centering
  \scriptsize
  \setlength{\tabcolsep}{1.2pt}%
  \caption{Full downstream performance (\%). Best scores within each bit width are in bold. Qwen3.5-35B-A3B and Kimi-Linear-48B-A3B are evaluated at 8 bits only. BA: BeyondAIME; MMLU: MMLU-Pro; USA: USACO; OJB: OJBench.}
  \label{tab:full-results}
  \settowidth{\apM}{FP8 group (E8M0)\ }%
  \apW=\dimexpr(\linewidth-\apM-8pt-30\tabcolsep)/16\relax
  \newcolumntype{N}{>{\raggedleft\arraybackslash}p{\apW}}%
  \begin{tabular}{@{}p{\apM}*{8}{N}@{\hspace{8pt}}*{8}{N}@{}}
    \toprule
     & \multicolumn{8}{c}{Qwen3.5-9B} & \multicolumn{8}{c}{Qwen3.8-Flash-Next} \\
    \cmidrule(lr){2-9}\cmidrule(l){10-17}
    Method & {\fontsize{6.2pt}{7pt}\selectfont AIME} & {\fontsize{6.2pt}{7pt}\selectfont BA} & {\fontsize{6.2pt}{7pt}\selectfont GPQA} & {\fontsize{6.2pt}{7pt}\selectfont LCB} & {\fontsize{6.2pt}{7pt}\selectfont HMMT} & {\fontsize{6.2pt}{7pt}\selectfont MMLU} & {\fontsize{6.2pt}{7pt}\selectfont USA} & {\fontsize{6.2pt}{7pt}\selectfont OJB} & {\fontsize{6.2pt}{7pt}\selectfont AIME} & {\fontsize{6.2pt}{7pt}\selectfont BA} & {\fontsize{6.2pt}{7pt}\selectfont GPQA} & {\fontsize{6.2pt}{7pt}\selectfont LCB} & {\fontsize{6.2pt}{7pt}\selectfont HMMT} & {\fontsize{6.2pt}{7pt}\selectfont MMLU} & {\fontsize{6.2pt}{7pt}\selectfont USA} & {\fontsize{6.2pt}{7pt}\selectfont OJB} \\
    \midrule
    FP32 & 87.9 & 63.3 & 81.3 & 64.1 & 77.7 & 82.9 & 24.0 & 19.6 & 98.5 & 85.1 & 91.4 & 86.8 & 98.3 & 88.0 & 76.0 & 54.8 \\
    \midrule
    BF16 & 72.1 & 33.7 & 66.2 & 49.6 & 51.5 & 81.8 & 16.7 & 10.7 & 96.8 & 78.8 & 90.2 & 84.0 & 97.2 & 88.2 & 67.6 & 41.7 \\
    \midrule
    \multicolumn{17}{@{}l}{\textit{8-bit methods}} \\
    \cellcolor{oursrow}\textbf{Ours} & \cellcolor{oursrow}\textbf{87.9} & \cellcolor{oursrow}\textbf{64.0} & \cellcolor{oursrow}\textbf{81.8} & \cellcolor{oursrow}\textbf{64.1} & \cellcolor{oursrow}\textbf{77.2} & \cellcolor{oursrow}\textbf{83.7} & \cellcolor{oursrow}\textbf{29.9} & \cellcolor{oursrow}\textbf{17.3} & \cellcolor{oursrow}\textbf{99.0} & \cellcolor{oursrow}\textbf{85.3} & \cellcolor{oursrow}\textbf{91.6} & \cellcolor{oursrow}\textbf{85.8} & \cellcolor{oursrow}\textbf{97.9} & \cellcolor{oursrow}87.5 & \cellcolor{oursrow}\textbf{77.0} & \cellcolor{oursrow}\textbf{53.0} \\
    FP8 per-tensor & 0.0 & 0.2 & 3.4 & 14.2 & 0.0 & 42.2 & 0.0 & 0.0 & 72.2 & 46.7 & 68.0 & 57.3 & 63.5 & 84.6 & 25.0 & 4.2 \\
    FP8 per-channel & 2.2 & 1.5 & 17.8 & 11.2 & 1.6 & 47.1 & 0.0 & 0.0 & 42.9 & 19.5 & 55.4 & 38.9 & 28.5 & 80.8 & 4.9 & 2.4 \\
    FP8 group (E8M0) & 0.0 & 0.1 & 2.2 & 5.9 & 0.0 & 31.9 & 0.0 & 0.0 & 82.5 & 52.8 & 76.8 & 71.5 & 73.9 & 85.7 & 37.7 & 15.5 \\
    FP8 group (FP32) & 0.0 & 0.1 & 3.5 & 4.1 & 0.2 & 24.5 & 0.0 & 0.0 & 42.4 & 17.2 & 52.9 & 39.4 & 28.6 & 79.2 & 6.9 & 1.2 \\
    INT8 per-tensor & 0.0 & 0.3 & 4.5 & 2.8 & 0.0 & 26.9 & 0.0 & 0.0 & 7.1 & 4.8 & 34.5 & 30.3 & 6.3 & 68.5 & 2.5 & 0.0 \\
    INT8 per-channel & 3.9 & 1.8 & 17.2 & 10.7 & 1.6 & 44.9 & 0.0 & 0.0 & 67.5 & 41.8 & 61.3 & 56.0 & 56.0 & 83.2 & 25.5 & 8.9 \\
    \addlinespace[1.5pt]
    KVQuant & 70.3 & 39.9 & 66.8 & 55.5 & 54.7 & 81.2 & 23.0 & 11.9 & 94.7 & 74.2 & 87.2 & 79.9 & 91.8 & \textbf{87.6} & 54.4 & 29.2 \\
    QuaRot & 11.4 & 7.4 & 38.2 & 35.6 & 12.1 & 72.8 & 2.9 & 3.6 & 88.3 & 61.0 & 84.7 & 72.0 & 81.3 & 87.1 & 42.6 & 20.8 \\
    TurboQuant & 67.9 & 43.2 & 62.0 & 42.2 & 51.1 & 80.8 & 10.3 & 5.4 & 95.4 & 78.4 & 89.9 & 84.7 & 96.6 & 87.1 & 67.2 & 41.1 \\
    \midrule
    \multicolumn{17}{@{}l}{\textit{6-bit methods}} \\
    \cellcolor{oursrow}\textbf{Ours} & \cellcolor{oursrow}\textbf{87.9} & \cellcolor{oursrow}\textbf{63.6} & \cellcolor{oursrow}\textbf{81.0} & \cellcolor{oursrow}\textbf{65.4} & \cellcolor{oursrow}\textbf{76.1} & \cellcolor{oursrow}\textbf{83.7} & \cellcolor{oursrow}\textbf{31.4} & \cellcolor{oursrow}\textbf{20.2} & \cellcolor{oursrow}\textbf{98.8} & \cellcolor{oursrow}\textbf{85.8} & \cellcolor{oursrow}\textbf{89.9} & \cellcolor{oursrow}\textbf{86.8} & \cellcolor{oursrow}\textbf{98.0} & \cellcolor{oursrow}\textbf{87.8} & \cellcolor{oursrow}\textbf{72.1} & \cellcolor{oursrow}\textbf{54.8} \\
    MXFP6 & 0.0 & 0.0 & 0.8 & 0.5 & 0.1 & 10.1 & 0.0 & 0.0 & 59.2 & 34.6 & 63.3 & 58.0 & 48.1 & 84.0 & 25.5 & 13.7 \\
    NVFP6 & 0.0 & 0.1 & 1.5 & 7.6 & 0.0 & 28.5 & 0.0 & 0.6 & 80.4 & 52.9 & 73.7 & 67.4 & 73.1 & 86.1 & 36.3 & 17.9 \\
    INT6 per-channel & 0.0 & 0.0 & 5.9 & 2.0 & 0.0 & 22.4 & 0.0 & 0.0 & 10.6 & 3.4 & 24.2 & 28.2 & 8.5 & 64.3 & 0.0 & 0.0 \\
    KVQuant & 1.5 & 0.7 & 10.1 & 15.5 & 1.4 & 39.7 & 0.0 & 0.0 & 55.6 & 32.7 & 54.5 & 48.3 & 44.9 & 81.8 & 17.2 & 3.0 \\
    QuaRot & 0.6 & 0.3 & 7.6 & 3.6 & 0.0 & 26.8 & 0.0 & 0.0 & 45.1 & 21.9 & 50.3 & 39.2 & 29.8 & 79.8 & 4.9 & 1.8 \\
    TurboQuant & 19.3 & 7.2 & 33.2 & 13.0 & 13.6 & 59.9 & 0.5 & 0.0 & 94.4 & 73.1 & 88.2 & 83.7 & 93.3 & 87.4 & 57.8 & 38.7 \\
    \midrule
    \multicolumn{17}{@{}l}{\textit{4-bit methods}} \\
    \cellcolor{oursrow}\textbf{Ours} & \cellcolor{oursrow}\textbf{85.0} & \cellcolor{oursrow}\textbf{60.1} & \cellcolor{oursrow}\textbf{80.0} & \cellcolor{oursrow}\textbf{61.3} & \cellcolor{oursrow}\textbf{72.0} & \cellcolor{oursrow}\textbf{83.0} & \cellcolor{oursrow}\textbf{25.5} & \cellcolor{oursrow}\textbf{16.7} & \cellcolor{oursrow}\textbf{92.5} & \cellcolor{oursrow}\textbf{72.6} & \cellcolor{oursrow}\textbf{90.6} & \cellcolor{oursrow}\textbf{79.9} & \cellcolor{oursrow}\textbf{86.6} & \cellcolor{oursrow}\textbf{87.7} & \cellcolor{oursrow}\textbf{58.8} & \cellcolor{oursrow}\textbf{40.5} \\
    MXFP4 & 0.0 & 0.0 & 3.7 & 0.3 & 0.0 & 8.5 & 0.0 & 0.0 & 1.7 & 2.8 & 25.4 & 22.6 & 2.1 & 65.9 & 0.0 & 0.0 \\
    NVFP4 & 0.0 & 0.0 & 1.0 & 0.0 & 0.0 & 1.6 & 0.0 & 0.0 & 0.1 & 0.0 & 4.5 & 14.5 & 0.0 & 44.5 & 0.0 & 0.0 \\
    INT4 per-channel & 0.0 & 0.0 & 5.7 & 0.0 & 0.0 & 6.8 & 0.0 & 0.0 & 0.0 & 0.0 & 6.2 & 2.0 & 0.1 & 21.3 & 0.0 & 0.0 \\
    KVQuant & 0.0 & 0.0 & 0.0 & 0.3 & 0.0 & 0.8 & 0.0 & 0.0 & 1.0 & 0.1 & 8.8 & 19.3 & 0.4 & 49.7 & 0.0 & 0.6 \\
    QuaRot & 0.0 & 0.0 & 0.5 & 0.0 & 0.0 & 1.7 & 0.0 & 0.0 & 0.0 & 0.1 & 3.5 & 13.0 & 0.0 & 39.6 & 0.0 & 0.0 \\
    TurboQuant & 3.5 & 1.4 & 26.8 & 11.5 & 1.5 & 48.7 & 0.0 & 0.0 & 79.3 & 49.8 & 74.9 & 62.8 & 69.0 & 85.6 & 34.8 & 15.5 \\
    \bottomrule
  \end{tabular}\par\vspace{1em}
  \sbox2{\begin{tabular}[t]{@{}p{\apM}*{8}{N}@{}}
    \toprule
    & \multicolumn{8}{c}{GLM-5.3-Flash} \\
    \cmidrule(l){2-9}
    Method & {\fontsize{6.2pt}{7pt}\selectfont AIME} & {\fontsize{6.2pt}{7pt}\selectfont BA} & {\fontsize{6.2pt}{7pt}\selectfont GPQA} & {\fontsize{6.2pt}{7pt}\selectfont LCB} & {\fontsize{6.2pt}{7pt}\selectfont HMMT} & {\fontsize{6.2pt}{7pt}\selectfont MMLU} & {\fontsize{6.2pt}{7pt}\selectfont USA} & {\fontsize{6.2pt}{7pt}\selectfont OJB} \\
    \midrule
    FP32 & 92.2 & 80.7 & 92.1 & 85.2 & 93.3 & 87.0 & 74.0 & 58.9 \\
    \midrule
    BF16 & 76.7 & 57.2 & 90.6 & 74.6 & 76.6 & 86.8 & 46.1 & 29.2 \\
    \midrule
    \multicolumn{9}{@{}l}{\textit{8-bit methods}} \\
    \cellcolor{oursrow}\textbf{Ours} & \cellcolor{oursrow}\textbf{91.1} & \cellcolor{oursrow}\textbf{83.2} & \cellcolor{oursrow}\textbf{91.2} & \cellcolor{oursrow}\textbf{87.5} & \cellcolor{oursrow}\textbf{94.7} & \cellcolor{oursrow}86.8 & \cellcolor{oursrow}\textbf{69.1} & \cellcolor{oursrow}\textbf{54.8} \\
    FP8 per-tensor & 19.9 & 5.7 & 48.5 & 30.3 & 11.4 & 72.7 & 1.5 & 0.6 \\
    FP8 per-channel & 31.9 & 14.2 & 66.3 & 36.4 & 23.6 & 80.8 & 1.0 & 1.2 \\
    FP8 group (E8M0) & 69.6 & 44.2 & 82.0 & 53.2 & 56.9 & 85.9 & 14.2 & 11.9 \\
    FP8 group (FP32) & 27.5 & 13.7 & 61.8 & 34.4 & 20.1 & 79.9 & 2.5 & 1.2 \\
    INT8 per-tensor & 28.2 & 10.0 & 57.1 & 29.0 & 14.9 & 79.3 & 1.5 & 1.2 \\
    INT8 per-channel & 87.4 & 66.8 & 90.4 & 67.4 & 81.9 & \textbf{86.9} & 34.8 & 19.0 \\
    \addlinespace[1.5pt]
    KVQuant & 87.9 & 72.8 & 90.6 & 83.0 & 89.3 & 86.8 & 54.4 & 36.9 \\
    QuaRot & 91.0 & 76.2 & 90.9 & 78.6 & 91.6 & 86.4 & 49.0 & 37.5 \\
    TurboQuant & 83.8 & 67.3 & 89.9 & 73.8 & 81.1 & 86.6 & 48.5 & 32.1 \\
    \midrule
    \multicolumn{9}{@{}l}{\textit{6-bit methods}} \\
    \cellcolor{oursrow}\textbf{Ours} & \cellcolor{oursrow}\textbf{92.9} & \cellcolor{oursrow}\textbf{82.7} & \cellcolor{oursrow}\textbf{91.2} & \cellcolor{oursrow}\textbf{86.3} & \cellcolor{oursrow}\textbf{94.7} & \cellcolor{oursrow}\textbf{87.2} & \cellcolor{oursrow}\textbf{69.6} & \cellcolor{oursrow}\textbf{50.0} \\
    MXFP6 & 71.4 & 44.8 & 82.7 & 53.4 & 57.2 & 85.7 & 13.7 & 13.7 \\
    NVFP6 & 39.2 & 17.1 & 67.8 & 36.9 & 22.0 & 83.1 & 3.4 & 1.2 \\
    INT6 per-channel & 48.2 & 26.1 & 74.6 & 36.4 & 35.9 & 84.4 & 2.5 & 1.2 \\
    KVQuant & 68.1 & 45.5 & 85.0 & 49.9 & 64.2 & 85.7 & 10.8 & 7.1 \\
    QuaRot & 39.7 & 19.6 & 72.2 & 36.1 & 29.2 & 83.4 & 3.4 & 1.8 \\
    TurboQuant & 88.1 & 66.8 & 88.0 & 66.2 & 79.5 & 86.5 & 30.9 & 17.3 \\
    \midrule
    \multicolumn{9}{@{}l}{\textit{4-bit methods}} \\
    \cellcolor{oursrow}\textbf{Ours} & \cellcolor{oursrow}\textbf{85.0} & \cellcolor{oursrow}\textbf{67.6} & \cellcolor{oursrow}\textbf{91.6} & \cellcolor{oursrow}\textbf{72.8} & \cellcolor{oursrow}\textbf{83.1} & \cellcolor{oursrow}\textbf{87.0} & \cellcolor{oursrow}\textbf{43.6} & \cellcolor{oursrow}\textbf{30.4} \\
    MXFP4 & 25.3 & 18.6 & 70.9 & 29.3 & 20.5 & 81.9 & 1.0 & 0.6 \\
    NVFP4 & 2.1 & 0.8 & 11.8 & 13.7 & 0.4 & 32.6 & 0.0 & 0.0 \\
    INT4 per-channel & 0.8 & 0.5 & 13.5 & 8.4 & 0.0 & 37.5 & 0.0 & 0.0 \\
    KVQuant & 5.0 & 1.5 & 30.6 & 16.5 & 1.7 & 41.6 & 0.0 & 0.0 \\
    QuaRot & 0.8 & 0.5 & 16.2 & 12.7 & 0.0 & 47.2 & 0.0 & 0.0 \\
    TurboQuant & 59.7 & 43.2 & 77.9 & 49.4 & 47.8 & 83.0 & 15.2 & 7.1 \\
    \bottomrule
  \end{tabular}}%
  \apLH=\dimexpr\ht2+\dp2\relax
  \newcommand{\rightbody}{%
    \toprule
    & \multicolumn{5}{c}{Qwen3.5-35B-A3B} \\
    \cmidrule(l){2-6}
    Method & {\fontsize{6.2pt}{7pt}\selectfont GSM} & {\fontsize{6.2pt}{7pt}\selectfont AIME} & {\fontsize{6.2pt}{7pt}\selectfont GPQA} & {\fontsize{6.2pt}{7pt}\selectfont LCB} & {\fontsize{6.2pt}{7pt}\selectfont MMLU} \\
    \midrule
    FP32 & 96.7 & 91.5 & 84.7 & 75.6 & 85.9 \\
    \midrule
    BF16 & 96.6 & 85.8 & 79.3 & 67.2 & 85.2 \\
    \midrule
    \multicolumn{6}{@{}l}{\textit{8-bit methods}} \\
    \cellcolor{oursrow}\textbf{Ours} & \cellcolor{oursrow}\textbf{96.5} & \cellcolor{oursrow}\textbf{91.0} & \cellcolor{oursrow}\textbf{83.9} & \cellcolor{oursrow}\textbf{76.1} & \cellcolor{oursrow}\textbf{85.8} \\
    FP8 per-tensor & 86.7 & 29.6 & 39.9 & 26.7 & 56.4 \\
    FP8 per-channel & 76.9 & 0.0 & 5.6 & 8.4 & 36.6 \\
    FP8 per-group & 67.2 & 0.0 & 1.5 & 2.3 & 24.9 \\
    MXFP8 & 80.6 & 29.6 & 36.4 & 29.8 & 45.5 \\
    INT8 per-tensor & 10.2 & 0.0 & 3.5 & 1.5 & 8.5 \\
    INT8 per-channel & 16.5 & 0.0 & 0.0 & 3.1 & 6.8 \\
    \addlinespace[1.5pt]
    KVQuant & 95.8 & 76.3 & 69.7 & 44.3 & 83.5 \\
    QuaRot & 80.7 & 32.9 & 36.4 & 15.3 & 56.0 \\
    TurboQuant & 82.9 & 10.4 & 22.7 & 15.3 & 52.7 \\
    \midrule[\heavyrulewidth]
    & \multicolumn{5}{c}{Kimi-Linear-48B-A3B} \\
    \cmidrule(l){2-6}
    FP32 & 92.1 & 67.5 & 70.3 & 41.4 & 72.4 \\
    \midrule
    BF16 & 92.1 & 64.3 & 68.1 & 41.0 & 64.0 \\
    \midrule
    \multicolumn{6}{@{}l}{\textit{8-bit methods}} \\
    \cellcolor{oursrow}\textbf{Ours} & \cellcolor{oursrow}92.0 & \cellcolor{oursrow}\textbf{68.3} & \cellcolor{oursrow}\textbf{69.8} & \cellcolor{oursrow}41.3 & \cellcolor{oursrow}72.1 \\
    FP8 per-tensor & 91.8 & 25.6 & 46.6 & 16.0 & 57.0 \\
    FP8 per-channel & 91.6 & 35.4 & 57.6 & 16.5 & 65.5 \\
    FP8 per-group & 91.7 & 25.3 & 47.6 & 15.8 & 63.5 \\
    MXFP8 & 91.4 & 43.5 & 50.3 & 23.4 & 67.0 \\
    INT8 per-tensor & 89.9 & 5.3 & 31.3 & 8.4 & 54.8 \\
    INT8 per-channel & 91.8 & 52.8 & 65.8 & 35.5 & 64.5 \\
    \addlinespace[1.5pt]
    KVQuant & 92.0 & 66.6 & 69.7 & \textbf{42.1} & 65.0 \\
    QuaRot & \textbf{92.6} & 63.3 & 69.2 & 36.6 & \textbf{73.1} \\
    TurboQuant & 92.3 & 57.5 & 69.2 & 41.2 & 69.0 \\
    \bottomrule
  }%
  \settowidth{\apRM}{INT8 per-channel}%
  \apRW=\dimexpr(\linewidth-\wd2-9pt-\apRM-10\tabcolsep)/5\relax
  \newcolumntype{M}{>{\raggedleft\arraybackslash}p{\apRW}}%
  \renewcommand{\arraystretch}{1}\sbox4{\begin{tabular}[t]{@{}p{\apRM}*{5}{M}@{}}\rightbody\end{tabular}}%
  \renewcommand{\arraystretch}{1.2}\sbox6{\begin{tabular}[t]{@{}p{\apRM}*{5}{M}@{}}\rightbody\end{tabular}}%
  \apS=\dimexpr 1pt+0.2pt*\numexpr\apLH-\ht4-\dp4\relax/\numexpr\ht6+\dp6-\ht4-\dp4\relax\relax
  \message{STRETCH=\strip@pt\apS}\edef\arraystretch{\strip@pt\apS}%
  \apRW=\dimexpr(\linewidth-\wd2-9pt-\apRM-10\tabcolsep)/5\relax
  \hbox to\linewidth{\usebox2\hfill\begin{tabular}[t]{@{}p{\apRM}*{5}{M}@{}}\rightbody\end{tabular}}
\end{table}
\makeatother

\end{document}